\documentclass[pdflatex,sn-mathphys-ay]{sn-jnl}

\usepackage{graphicx}%
\usepackage{multicol,multirow}
\usepackage{amsmath,amssymb,amsfonts}%
\usepackage{amsthm}%
\usepackage{mathrsfs}%
\usepackage[title]{appendix}%
\usepackage{xcolor}%
\usepackage{textcomp}%
\usepackage{manyfoot}%
\usepackage{booktabs}%
\usepackage{algorithm}%
\usepackage{algorithmicx}%
\usepackage{algpseudocode}%
\usepackage{listings}%

\usepackage{tabularx}
\usepackage{soul}
\usepackage[utf8]{inputenc}
\usepackage[T1]{fontenc}  
\usepackage{xstring}
\usepackage{array}
\usepackage{xcolor}
\usepackage{enumitem}
\usepackage{siunitx}
\usepackage{booktabs}%
\usepackage{color}
\usepackage{xargs}                      
\usepackage[many]{tcolorbox}
\usepackage{lineno}
\usepackage{tikz}
\usetikzlibrary{positioning, arrows.meta, bending}

\tcbset{
    speechbox/.style={
        colback=gray!5,
        colframe=black!75,
        boxrule=0.5pt,
        arc=2mm,
        left=4pt,
        right=4pt,
        top=4pt,
        bottom=4pt,
        width=0.95\textwidth,
        fonttitle=\bfseries
    },
        templatebox/.style={
        colback=white,
        colframe=black!75,
        boxrule=0.5pt,
        arc=2mm,
        fontupper=\ttfamily\small,
        left=4pt,
        right=4pt,
        top=4pt,
        bottom=4pt
    }
}

\makeatletter
\newcommand*\iftodonotes{\if@todonotes@disabled\expandafter\@secondoftwo\else\expandafter\@firstoftwo\fi}  
\makeatother

\newcommandx{\finalpass}[2][1=]{\todo[linecolor=red,backgroundcolor=red!50,bordercolor=red,#1]{General check $\rightarrow$ #2}}
\newcommandx{\add}[2][1=]{\todo[color=red,#1]{Add $\rightarrow$ #2}}
\newcommandx{\improve}[2][1=]{\todo[linecolor=OliveGreen,backgroundcolor=OliveGreen!25,bordercolor=OliveGreen,#1]{Improve $\rightarrow$ #2}}
\newcommandx{\change}[2][1=]{\todo[linecolor=blue,backgroundcolor=blue!25,bordercolor=blue,#1]{Change $\rightarrow$ #2}}

\newcommandx{\Finalpass}[2][1=]{\todo[inline,linecolor=red,backgroundcolor=red!50,bordercolor=red,#1]{General check $\rightarrow$ #2}}
\newcommandx{\Add}[2][1=]{\todo[inline,color=red,#1]{Add $\rightarrow$ #2}}
\newcommandx{\Improve}[2][1=]{\todo[inline,linecolor=OliveGreen,backgroundcolor=OliveGreen!25,bordercolor=OliveGreen,#1]{Improve $\rightarrow$ #2}}
\newcommandx{\Change}[2][1=]{\todo[inline,linecolor=blue,backgroundcolor=blue!25,bordercolor=blue,#1]{Change $\rightarrow$ #2}}

\AtBeginEnvironment{appendices}{%
}

\theoremstyle{thmstyleone}%
\theoremstyle{thmstyletwo}%

\theoremstyle{thmstylethree}%
\newtheorem{definition}{Definition}%

\begin{document}

\title[EURO-5K]{EURO-5K: When Does Domain Pretraining Matter? Benchmarking Transformers for EU Reporting Obligation Extraction}


\author*[1]{\fnm{Marios} \sur{Koniaris}}\email{mkoniari@central.ntua.gr}
\author[1]{\fnm{Vasileios} \sur{Kotronis}}\email{el21432@mail.ntua.gr}
\author[2]{\fnm{Eugenia} \sur{Giannini}}\email{giannini@mail.ntua.gr}
\author[1]{\fnm{Panayiotis} \sur{Tsanakas}}\email{panag@cs.ntua.gr}
 
\affil*[1]{\orgdiv{Division of Computer Science, School of Electrical and Computer Engineering}, \orgname{National Technical University of Athens}, \orgaddress{\street{Iroon Polytechniou 9}, \city{Zographou Campus}, \postcode{15780}, \state{Athens}, \country{Greece}}}

\affil[2]{\orgdiv{Department of Humanities Social Sciences and Law, School of Applied Mathematical and Physical Sciences,}, \orgname{National Technical University of Athens}, \orgaddress{\street{Iroon Polytechniou 9}, \city{Zographou Campus}, \postcode{15780}, \state{Athens}, \country{Greece}}}

\abstract{Extracting reporting obligations from EU legislation is critical for assessing and reducing regulatory reporting burden. However, distinguishing reporting requirements from structurally similar provisions requires specialised legal understanding. Current legal NLP methods lack specialised datasets with clear guidelines and comparative evaluation of extraction paradigms and domain adaptation strategies.
We curate EURO-5K, a corpus of sentence-level reporting obligations and challenging negative examples from 136 EU legislative acts. On this dataset, we train and compare discriminative token-classification models (BERT-style) and generative span-extraction models (LLMs), evaluating both full fine-tuning and parameter-efficient QLoRA against baselines (pattern and dependency-based extraction, few-shot prompting).
Results show that fully fine-tuned generic and legal BERT models achieve similar performance (0.89 F1), while fine-tuned LLMs match encoder accuracy for sentence-level extraction. Legal pretraining offers only small gains for generative models. In contrast, it is clearly beneficial when adaptation capacity is constrained, as parameter-efficient tuning of Legal-BERT outperforms its generic counterpart. Learning curve analysis demonstrates that legal pretraining accelerates early learning with minimal data. All approaches converge around 3K samples with diminishing returns thereafter, validating dataset sufficiency. Cross-dataset evaluation on two external regulatory corpora shows that our models behave as specialised reporting obligation extractors rather than generic regulatory classifiers.
We release EURO-5K, trained models, and an interactive demo with explainability visualizations and structured RDF export. These demonstrate that both paradigms and parameter-efficient training provide practical tools for regulatory compliance automation.}


\maketitle

\section{Introduction} \label{sec:intro}

The European Union maintains approximately 180K legal acts spanning diverse policy domains. Finding specific obligations within this corpus is challenging, as not all duties require reporting. For instance, in the context of water safety, a behavioral obligation may mandate that \textit{“Member States shall ensure that the supply, treatment and distribution of water intended for human consumption is subject to a risk-based approach”} (Directive (EU) 2020/2184, Art. 7, CELEX:32020L2184). While legally binding, this establishes a substantive duty of conduct rather than a data transmission requirement. In contrast, a reporting obligation in the same Directive dictates that Member States \textit{“Member States shall notify the Commission of those rules and of those measures and shall notify it of any subsequent amendment affecting them.”} (Art. 23). EU policy analysis distinguishes~\citep{MARCUS2025} three obligation types: i) reporting obligations (e.g., submit data to authorities), ii) behavioral obligations (e.g., conduct activities), and iii) disclosure obligations (e.g., make information public).

Consider the 2025 Omnibus simplification package, enacted as Directive (EU) 2025/794 (CELEX:32025L0794). Manual analysis of three sustainability frameworks identified overlapping obligations, enabling removal of 80\% of companies from reporting scope, projected to save €4.4 billion annually~\citep{EC2025Omnibus}. This demonstrates both the scale of reporting burden and the potential for systematic optimization. Beyond burden reduction, reporting obligations constitute the primary feedback mechanism through which implementing institutions provide compliance data to legislative bodies, enabling assessment of regulatory effectiveness in practice. Automated extraction could extend this approach across the full EU legislative corpus, addressing challenges faced by initiatives like the Commission's 25\% burden reduction target and the Reduce the Reporting and Monitoring Burden (RRMV) program~\citep{rrmv}. 

Prior computational work targets broad categories without distinguishing obligation types. We focus on reporting obligations, requirements that underpin regulatory oversight and shape operational practices across sectors. Extracting reporting obligations from EU legislation presents challenges beyond general text classification. Legislative provisions exhibit deeply nested syntactic structures where obligations span multiple clauses and cross-references. Deontic markers like \textit{shall submit} appear not just in reporting obligations but also in procedural provisions and permissions. Distinguishing these requires deeper understanding of the regulatory context to identify who reports to whom and for what supervisory purpose. 

Manual annotation requires legal expertise to interpret whether a provision mandates upward information flow to authorities or coordinates peer-level communication between regulatory bodies. This expertise requirement limits scalability as analyzing thousands of legislative acts manually is prohibitively time-consuming and expensive. Automation through supervised learning could address this challenge. Yet the field lacks three prerequisites: specialized annotated datasets, trained models for reporting obligation extraction, and established methodologies for sentence-level extraction from regulatory text.

To enable supervised learning for this specialized task, we curated EURO-5K, a dataset of 5,253 sentences from EU legislative documents. Annotation required developing clear protocols to operationalize the reporting obligation concept, systematically sampling challenging negative examples, and validating quality through inter-annotator agreement. The resulting dataset provides the foundation for training and evaluating extraction models. We train models on EURO-5K, comparing generic (BERT-base, Mistral-7B, Llama-3.1-8B) and legal-domain (Legal-BERT, Saul-7B) transformers using discriminative token classification (BERT) and generative approaches (LLMs). We evaluate full fine-tuning and parameter-efficient LoRA for BERT models, and QLoRA for LLMs.

Our evaluation demonstrates that both discriminative and generative paradigms achieve comparable performance for specialized legal extraction. Legal pretraining provides modest but consistent benefits for discriminative methods and negligible benefits for generative approaches at 7B scale. Data efficiency analysis indicates that while legal pretraining accelerates early-stage learning, performance across all architectures converges at approximately 3,000 samples, establishing a clear threshold for diminishing returns in domain-specific annotation. Explainability analysis reveals models emphasize institutional actors and evaluate context rather than keywords. Finally, cross-dataset validation confirms that our trained models extract reporting obligations specifically rather than classifying regulatory statements broadly.

The contributions of this paper are as follows: 

\textbf{(i)} We build and release EURO-5K, the largest annotated corpus for reporting obligation extraction (5,253 sentences, 1,751 positives, 136 EU legislative documents), with a principled annotation protocol distinguishing reporting obligations from behaviourally and disclosure-related confounds, including 532 hard negatives targeting challenging boundary cases.

\textbf{(ii)} We train and compare generic and legal-domain transformers across discriminative and generative paradigms with different training methods, revealing that domain adaptation value depends on both paradigm and training strategy. Learning curve analysis reveals minimum training requirements and validates dataset sufficiency.

\textbf{(iii)} We systematically compare full fine-tuning and parameter-efficient methods (LoRA for BERT, QLoRA for LLMs) across discriminative and generative paradigms, revealing accuracy-efficiency trade-offs for specialized legal extraction

\textbf{(iv)} Cross-dataset evaluation on two external corpora validates specialized learning along complementary axes: \emph{specificity} (correctly rejecting general regulatory statements) and \emph{sensitivity} (88.7--90.3\% zero-shot recall on out-of-domain financial reporting obligations).

\textbf{(v)} We provide comprehensive statistical significance testing using Welch's t-test and bootstrap methods, confirming that domain adaptation effects are statistically non-significant while paradigm parity is statistically validated.

\textbf{(vi)} We demonstrate deployment readiness through an interactive web interface integrating model predictions, explainability visualizations, and RRMV-compliant RDF export for regulatory knowledge bases.

The remainder of this paper is organized as follows: Section~\ref{sec:related} reviews related work on legal information extraction and transformer architectures. Section~\ref{sec:method} presents our dataset curation methodology. Section~\ref{sec:experiments} presents experimental configurations. Section~\ref{sec:results} reports extraction performance, cross-dataset generalization, and explainability findings. Section~\ref{sec:applications} demonstrates practical deployment. Finally, we draw our conclusions and present future work aspects in Section~\ref{sec:conclusion}.

\section{Related Work} \label{sec:related}

Legal information extraction has evolved from rule-based systems to deep learning and, more recently, generative AI. Our work intersects legal domain language models, regulatory obligation extraction, information extraction paradigms, and parameter-efficient training.

\subsection{Legal NLP and Domain Adaptation}

Legal text presents distinct challenges for NLP through specialized vocabulary, formal syntax, and domain-specific semantics. Continued pre-training on legal corpora tends to outperform generic models~\citep{chalkidislegal}, with performance gains scaling with task specificity~\citep{Zheng2021}. However, questions remain whether gains justify computation costs~\citep{hendrycks2020measuring}. Parameter-efficient alternatives have been explored for resource-constrained scenarios~\citep{HerrewijnenC23}.

For regulatory obligation extraction, the Institutional Grammar Tool~\citep{crawford1995grammar,ostrom2009} provides a formal framework distinguishing obligations from other regulatory statements through Attribute, Deontic, and Aim components. Deontic classification identifies obligations, permissions, and prohibitions~\citep{liga2022deontic}. Evaluation on EU regulatory statements achieved high accuracy~\citep{Brandsma2025ComputationalIOB}, while contract analysis explored deontic modality classification~\citep{Graham2023}. Extraction of judicial interpretative formulas from CJEU case law demonstrated that fine-tuned BERT models achieve performance comparable to generative models while offering advantages in stability and reproducibility~\citep{Grundler2025}. Low-resource approaches have investigated limited labeled data scenarios~\citep{chakravarthy2025ner}.

Beyond binary classification, structured extraction approaches have emerged using LLM prompt engineering~\citep{ZinSB24} and few-shot methods combining dependency parsing with classification~\citep{Pawar2023}. We focus on a specialized subset reporting obligations requiring entities to submit data to authorities. and move beyond binary classification to sentence-level extraction for this policy-critical subset.

Closely related, \citet{corazza2025} detect reporting requests in EU legislation with a hybrid approach, comparing BERT variants and in-context LLMs on 991 paragraphs, while \citet{dalpont2025} extract broader deontic obligations from the GDPR, DSA, and AI Act via an LLM pipeline with knowledge-graph output. We complement these with a larger annotated corpus (EURO-5K) and the first evaluation of fine-tuned LLMs (QLoRA) for reporting obligation extraction.

\subsection{Information Extraction Paradigms}

Token classification has dominated legal entity recognition and span extraction, with applications in deontic sentence classification~\citep{liga2022deontic,Minkova2023}. Fine-tuning generative models for legal classification has shown substantial advantages over discriminative approaches~\citep{ligarobaldo2023gpt3}, though performance depends on task characteristics. Alternative architectures have achieved strong results on sentence-level tasks~\citep{Pennisi2023NOMOS}.

Evaluations across paradigms~\citep{GrundlerLMLGST24} reveal complementary strengths, as discriminative models excel at detection with adequate training data, while generative models handle sparse multi-class scenarios better. Zero-shot LLM approaches for legal extraction have shown limitations, with regex-based methods achieving better precision in some cases~\citep{Molinari2025}. Zero-shot performance varies across legal document types~\citep{Savelka2023}, with regulatory text posing particular challenges. Few-shot learning~\citep{TerronMR23} and query-based methods~\citep{ZinNSSN23} have been explored to improve performance, underscoring the importance of task-specific fine-tuning for specialized legal extraction.

We provide the first systematic comparison for sentence-level reporting obligation extraction from EU regulations, evaluating both fine-tuned discriminative and generative models alongside few-shot prompting baselines, with multi-seed statistical validation and cross-dataset generalisation testing.

\subsection{Large Language Models for Legal Tasks}

LLM capabilities in legal applications have been extensively evaluated. \citep{Katz2023} demonstrated that models can apply complex legal principles on standardized legal assessments, though challenges remain in nuanced reasoning tasks. Comprehensive benchmarks covering diverse legal reasoning categories revealed significant performance variability ~\citep{Guha2023legalbench}. Legal models have been developed through continued pre-training on legal corpora ~\citep{saul2024}. These capabilities come with limitations. Analysis of legal hallucinations, where models generate outputs deviating from facts, found high error rates influenced by jurisdiction and case characteristics~\citep{Dahl2024}, emphasizing the need for human oversight.

\subsection{Parameter-Efficient Training}

Growing computational demands have prompted parameter-efficient alternatives. Quantized Low-Rank Adaptation (QLoRA) ~\citep{dettmers2024qlora} combines 4-bit quantization with low-rank adapter training, updating under 1\% of parameters while enabling consumer hardware deployment. Applications to legal tasks demonstrated training time reductions while maintaining performance~\citep{ChangC23}. Performance analysis confirms QLoRA's memory savings come at the cost of increased computation time due to dequantization overhead~\citep{Hanindhito2025}.

Parameter-efficient approaches for legal domain adaptation remain underexplored. We evaluate FFT and parameter-efficient adaptation (LoRA for BERT, QLoRA for LLMs) across discriminative and generative approaches, comparing domain adaptation effects across training strategies.

\section{Methodology} \label{sec:method}

We formulate the extraction of reporting obligations from EU legislation as a supervised learning task. This section defines the problem and task, then describes the dataset used for training and evaluation.

\subsection{Task Definition}

Given a legislative text $T$ comprising sentences $\{s_1, s_2, ..., s_n\}$, our goal is to identify the subset $S \subseteq \{s_1, ..., s_n\}$ where each sentence $s_i \in S$ contains a reporting obligation. Formally:

\begin{equation}
S = \{s_i \in T \mid \text{isReportingObligation}(s_i) = \text{true}\}
\end{equation}

where $\text{isReportingObligation}(s_i)$ evaluates whether sentence $s_i$ mandates information submission from an entity to a regulatory authority for supervisory purposes. This task is challenging due to the substantial lexical and structural overlap between reporting obligations and other regulatory statements, requiring the combined modeling of linguistic cues and contextual semantics. 

Legal obligations are duties mandated by statute, regulation, or contract, enforceable within a jurisdiction~\citep{FIBO}. Reporting obligations constitute a subclass requiring data submission to authorities, distinct from contractual or behavioral obligations. Following the analytical framework established by ~\citep{MARCUS2025}, we distinguish Reporting Obligations (administrative requirements for information transmission) from Behavioural Obligations (substantive duties of conduct) and Disclosure Obligations (public transparency requirements). This distinction ensures our extraction task targets the specific transmission of data to regulatory authorities, rather than broader corporate statements. 

\begin{definition}[Reporting Obligation]\label{def:ro}
\normalsize
A reporting obligation is a mandatory legal requirement for a regulated subject to submit specific information to a regulatory or oversight authority for purposes of supervision, enforcement, or regulatory coordination. It encompasses both \textit{upward information submission} (operators or Member States reporting to the Commission or oversight bodies) and \textit{cross-border supervisory reporting} (home authority reporting to host authority for oversight purposes), spanning institutional reporting (regulatory agencies submitting to oversight bodies) and commercial compliance reporting (operators to competent authorities). It excludes: (1) behavioural obligations (conduct requirements), (2) disclosure obligations (subject publishes publicly), and (3) non-supervisory procedural coordination between peer authorities.
\end{definition}

This definition operationalises the European Commission's policy definition of regulatory reporting~\citep{EC2019casestudy}, refining it for sentence-level annotation: we add a supervisory-purpose criterion and explicitly exclude public disclosure and peer-level coordination, while relaxing the periodicity requirement to capture event-triggered obligations.

Consider for example the sentence: \textit{"The annual report to be submitted by Member States to the Commission shall include at least the following information..."}. While this mentions submission, the main verb \textit{"shall include"} specifies content requirements, not the duty to submit. The submission is background context; the primary obligation is what content must be included. Such content specifications assume the reporting obligation is established elsewhere. Documentation requirements enabling oversight, compound sentences mixing obligation with content, wrong-direction communications (authority to public), and content specifications assuming obligations elsewhere should not be mistakenly identified as reporting obligations.

The challenge lies in distinguishing reporting obligations from structurally similar statements using linguistic markers and contextual semantics. We address this through span detection at sentence level, where the goal is identifying complete obligation statements. This differs from document-level classification by requiring precise boundary identification. 

\subsection{Dataset}

The Annotation of Reporting Obligations in EU Legislation Dataset~\citep{aroldjrc} (AROLD) contains XML files of EU legislative documents with manual annotations by legal experts from the Joint Research Centre (JRC). Each document contains annotations marking reporting requirements. After resolving minor XML formatting issues, the raw dataset contained 30,432 positive annotations from 136 documents across 187,829 total sentences.

\begin{figure}[h]
\centering
\resizebox{\textwidth}{!}{
\begin{tikzpicture}[
  node distance=1.5cm,
  box/.style = {rectangle, draw=black!50, rounded corners=4pt, 
                text width=3.2cm, minimum height=2.5cm, align=center, 
                font=\scriptsize, fill=white, line width=0.5pt},
  headerbox/.style = {box, fill=white},
  processbox/.style = {box, fill=white},
  resultbox/.style = {box, fill=gray!2},
  arrow/.style={->, thick, >=stealth, color=black!60}
]

\node[headerbox] (raw) {
  \textbf{\normalsize Raw Data} \\[0.3em]
  30,432 annotations \\[0.3em]  
  {\scriptsize Structural Noise} \\
  {\scriptsize Multi-sentences } \\
  {\scriptsize Errors}\\
};

\node[processbox, right=of raw] (curation) {
  \centering    
  \textbf{\normalsize Curation} \\[0.3em]  
    Resegmentation  \\
    Rules filtering \\
    LLM review \\
    Human validation\\
};

\node[processbox, right=of curation] (iter) {
  \textbf{\normalsize Refinement} \\[0.3em]  
  Bert Model assisted \\
  FP discovery \\
    3 rounds \\  
};

\node[resultbox, right=of iter] (final) {
  \textbf{\normalsize EURO-5K} \\[0.3em]
  5,253 examples \\
  1,751 positive \\
  3,502 negative \\[0.2em] 
};

\coordinate (A) at (iter.south east);
\coordinate (B) at ([yshift=-0.7cm]iter.south);
\coordinate (C) at (iter.south west);

\draw[arrow, dotted] (raw) -- (curation);
\draw[arrow, dotted] (curation) -- (iter);
\draw[arrow, dotted] (iter) -- (final);

\draw[arrow, dashed, color=black!50, thick] 
      (A) .. controls ++(0.3,-0.5) and ++(0.3,0) .. 
      (B) .. controls ++(-0.3,0) and ++(0.3,-0.5) .. 
      (C);

 \node[below=0.1cm of B, font=\tiny, color=black!60] {iterative refinement};
\end{tikzpicture}
}
\caption{Data curation pipeline transforming raw AROLD annotations to final dataset. The process addresses structural noise, multi-sentence segmentation, and misclassifications through rule-based filtering, legal-aware resegmentation, LLM review with human validation, and three rounds of iterative model-driven refinement.}
\label{fig:curation_pipeline}
\end{figure}
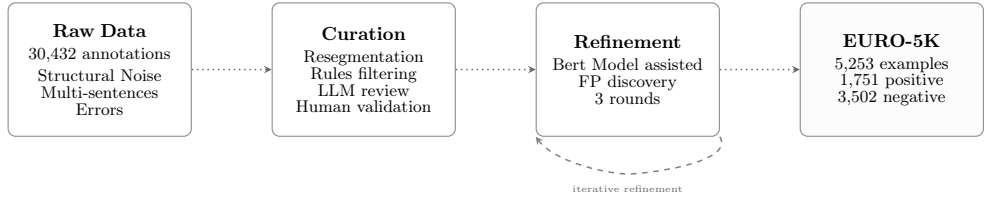

Initial analysis revealed several systematic problems. First, the dataset contained structural noise as several form labels or section headers were identified as obligations. Second, many annotations spanned multiple sentences making it unclear which specific sentence contained the obligation; approximately 35\% of annotations were multi-sentence. Third, some annotations were misclassified under Definition~\ref{def:ro}, including behavioural requirements or disclosure mandates rather than reporting obligations. These issues necessitated a curation process before model training. Figure~\ref{fig:curation_pipeline} shows the multi-stage validation process we employed.

To systematically address these issues, we applied five criteria distinguishing reporting obligations from related regulatory statements. A sentence is \textit{annotated as a reporting obligation} if it: 
\begin{enumerate}[label=(\roman*), itemsep=0.3em, topsep=0pt]
    \item contains a reporting action (e.g., submit, notify);
    \item uses mandatory language (e.g., shall, must);
    \item targets a regulatory or oversight authority;
    \item concerns information submission rather than operational conduct; and
    \item primarily expresses an obligation rather than specifying the content of a report.
\end{enumerate}

We applied a pipeline combining automated and human validation. The pipeline included rule-based filtering using reporting verbs and deontic modals, legal-aware sentence resegmentation for multi-sentence chunks, LLM-assisted review (Claude Sonnet 4) with a structured prompt encoding the five annotation criteria, and dual-blind human validation (achieving agreement, $\kappa = 0.613$)~\citep{artstein2008,krippendorff2018}. We then performed three rounds of iterative model-driven refinement. In each round, a BERT model was trained on the current annotations and used to surface potential reporting obligations from the negative set. These candidates were then subjected to the same LLM-assisted and human validation protocol. Following curation completion, this intermediary model was discarded and all evaluation models (Section \ref{sec:experiments}) were trained from scratch on the finalized dataset to prevent optimization bias.

Resulting dataset, EU-Reporting-Obligations-5K (EURO-5K), consists of 5,253 sentence-level examples, of which 1,751 are reporting obligations (positives) and 3,502 are non reporting obligations (negatives), maintaining a 1:2 ratio. Negative examples include 532 \textit{hard negatives} (10.3\%) which represent challenging cases like behavioral requirements and procedural coordination that prevent superficial pattern learning ~\citep{swayamdipta2020dataset}.

\begin{table}[h]
\caption{EURO-5K dataset structure and metadata fields.}
\label{tab:euro5k_metadata}
\centering
\begin{tabular}{@{}llp{6.5cm}@{}}
\toprule
\textbf{Field} & \textbf{Type} & \textbf{Description} \\
\midrule
id & Integer & Unique sentence identifier  \\
document\_id & String & Source document identifier from AROLD corpus \\
sentence\_num & Integer & Sentence position within source document \\
sub\_sentence\_idx & Integer & Sub-sentence index (for resegmented multi-sentence annotations) \\
sentence & String & Legislative sentence text in English \\
previous\_sentence & String & Preceding sentence for context resolution \\
next\_sentence & String & Following sentence for context resolution \\
has\_reporting\_obligation & Binary & Ground truth label \\
source & Categorical & \textit{hard\_negative} (challenging case), or inferred positive \\
document\_url & URL & Source document EUR-Lex URL \\
document\_title & String & Official document title \\
\bottomrule
\end{tabular}
\end{table}

Table~\ref{tab:euro5k_metadata} describes the dataset structure. Sentence lengths in EURO-5K range from 5 to 347 words (mean: 37±19, median: 33). The shortest sentences use pronouns referencing prior context (e.g., \textit{``They shall immediately inform the Commission thereof''}), while the longest span complex multi-clause provisions. Negative examples are slightly longer on average (38±18 words) than positives (35±19), as regulatory statements without reporting obligations often contain procedural or definitional complexity. Context fields (previous/next sentences) support pronoun resolution and cross-reference understanding. The 532 hard negatives (10.3\%) represent challenging cases with obligation-like language that prevent superficial pattern learning.

Examples are split document-level into train (70\%), validation (15\%), and test (15\%) sets to prevent data leakage. To ensure balanced representation across all splits, we employ stratified sampling by source type (positive, hard negative), maintaining proportional representation of each category across training, validation, and test sets. This prevents under representation of challenging cases that could compromise model evaluation. 

\section{Experimental Setup} \label{sec:experiments}

We compare generic and legal-domain transformers across two extraction paradigms: token classification (BERT-base, Legal-BERT) and generative text extraction (Mistral-7B, Saul-7B, Llama-3.1-8B). This design isolates domain adaptation effects across paradigms and training strategies (full fine-tuning vs parameter-efficient methods).

\begin{figure}[h] 
\centering
\includegraphics[width=0.95\columnwidth]{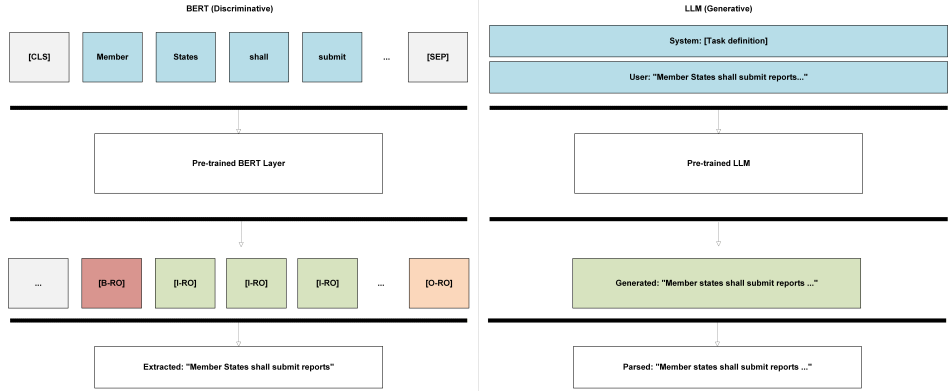}
\caption{Comparison of discriminative token classification and generative span extraction paradigms. Both process the same legislative sentence but employ fundamentally different architectures to identify reporting obligations.}
\label{fig:paradigm_comparison}
\end{figure}

Figure~\ref{fig:paradigm_comparison} illustrates the fundamental difference between the two extraction paradigms. Discriminative models (BERT-based) assign BIO labels to each token and group consecutive tags to extract obligation spans. Generative models (LLM-based) produce the obligation text directly through conditional generation, outputting either the extracted sentence or "None" for negative cases.

\subsection{Discriminative Models}

We evaluate BERT-base~\citep{devlin2019bert} and its legal-adapted variant Legal-BERT~\citep{chalkidislegal}.
We train with two strategies: Full Fine-Tuning (FFT) updating all 110M parameters, and LoRA~\citep{hu2021lora} updating 2.5M adapter parameters (2.3\% of model). 

For BERT models, we optimized hyperparameters, for each model, using Optuna \citep{akiba2019optuna,dodge2020finetuning}, a Bayesian hyperparameter optimization framework. Each model underwent 50 trials with 5-seed evaluation per trial, using Tree-structured Parzen Estimator (TPE) sampling to efficiently explore the hyperparameter space ~\citep{tpe2025}. The search space included learning rate, batch size, epochs, warmup ratio, weight decay, dropout rates, freeze layers, and LLRD decay. Table~\ref{tab:training_config} summarizes the resulting optimal configurations.

A token classification head predicts labels for each input token. The BIO scheme uses three labels: B-RO (beginning of reporting obligation), I-RO (inside), and O (outside), enabling precise boundary identification at token level.  Token-level BIO predictions convert to text spans by grouping consecutive B-RO and I-RO tokens, with subword tokens aligned back to original word boundaries.

\begin{table}[ht]
\centering
\small 
\caption{Training hyperparameters for all approaches. BERT models optimized via Optuna; LLM hyperparameters follow standard practice. Adapter parameters ($r$=rank, $\alpha$=scaling) apply to LoRA (BERT) and QLoRA (LLMs with 4-bit quantization). All configurations use AdamW optimizer, linear warmup, and load-best-model-at-end strategy.}
\label{tab:training_config}
\setlength{\tabcolsep}{4pt} 
\begin{tabular}{@{}lccccc@{}}
\toprule
\textbf{Approach} & \textbf{Base Model} & \textbf{LR} & \textbf{Batch} & \textbf{Epochs} & \textbf{Adapters} \\
\midrule
BERT-base (FFT) & bert-base & 7.4e-5 & 16 & 20 & -- \\
BERT-base (LoRA) & bert-base & 3.8e-5 & 16 & 18 & $r$=24, $\alpha$=96 \\
Legal-BERT (FFT) & Legal-BERT & 5.8e-5 & 16 & 17 & -- \\
Legal-BERT (LoRA) & Legal-BERT & 4.0e-5 & 16 & 17 & $r$=24, $\alpha$=76 \\
\midrule
Llama-QLoRA & Llama-3.1-8B & 1.0e-4 & 64 & 9 & $r$=16, $\alpha$=32 \\
Mistral-QLoRA & Mistral-7B & 1.0e-4 & 64 & 9 & $r$=16, $\alpha$=32 \\
Saul-QLoRA & Saul-7B & 1.0e-4 & 64 & 9 & $r$=16, $\alpha$=32 \\
\bottomrule
\end{tabular}
\end{table}

\subsection{Generative Models}

We employ Llama-3.1-8B~\citep{dubey2024llama}, Mistral-7B~\citep{jiang2023mistral} and Saul-7B~\citep{saul2024}. Llama-3.1 features 128k-token context window, while Mistral uses grouped query attention for efficiency. Saul-7B extends Mistral through continued pre-training on 30B tokens of legal text, testing whether legal domain adaptation benefits generative extraction. QLoRA~\citep{dettmers2024qlora,hu2021lora}  applies 4-bit quantization  with LoRA adapters to query and value projections, training only ~0.5\% of parameters. For LLM models, we use standard hyperparameters for consistent comparison (Table~\ref{tab:training_config}).

For generative models the task is conditional text generation. The instruction-following format uses: task description plus legislative text as input, with expected output being the extracted obligation text or "None" if absent. We extract the obligation text from generated outputs and parse 'None' responses as negatives. 

\subsection{Baseline Methods}\label{subsec:baseline}

\textit{RegEx/Keyword Matching.} To establish baseline performance we use rule-based pattern matching that combines deontic modals (e.g., shall, must) with reporting verbs (e.g., submit, report). Sentences containing these patterns become candidate obligations.

\noindent\textit{Dependency Parsing.} We use dependency parsing to identify obligations through grammatical structure. The method targets sentences where deontic modals (e.g., shall, must) govern reporting verbs (e.g., submit, report) with prepositional phrases indicating regulatory recipients. This tests whether explicit grammatical rules can capture obligation structure without machine learning.

\noindent\textit{Few-Shot Prompting.} We prompt Llama-3.1-8B Instruct model with task instructions and 5 labeled examples (3 positive, 2 negative) from the training set before each test instance. We use temperature $T = 0$ and disable sampling for deterministic outputs. This evaluates the model's ability to extract obligations from in-context examples alone, without parameter updates, directly comparing prompting versus fine-tuning strategies~\citep{GraySOA23}.

\subsection{Evaluation Framework}

We evaluate all approaches using standard metrics i.e., Precision, Recall, and F1-score. A prediction is correct under exact match if it precisely matches gold-standard boundaries. These span-level metrics reflect ability to identify complete, usable obligation statements. For BERT, we train with 5 random seeds and report mean±standard deviation. For LLMs, trained with single seeds due to computational constraints, we use bootstrap resampling (n=1,000 iterations) ~\citep{dror-stats} to estimate confidence intervals:

\begin{equation}
CI_{95\%} = [q_{0.025}(F_1^*), q_{0.975}(F_1^*)]
\end{equation}

where $F_1^*$ represents the empirical distribution of resampled F1 scores and $q_{\alpha}$ denotes the $\alpha$-quantile. Bootstrap standard deviations are comparable to BERT's multi-seed variance, enabling fair statistical comparison.

Statistical significance of performance differences is assessed via Welch's t-test for multi-seed BERT comparisons (accounting for potentially unequal variances across seeds) and bootstrap paired testing (10,000 iterations with stratified resampling) for cross-model comparisons, following established practices in NLP evaluation~\citep{dror-stats,reimers2018reporting}.

To ensure fair comparison all approaches evaluate on identical held-out test sets and post-processing standardizes outputs to span-level format. We report energy consumption and trainable parameters to assess sustainability trade-offs. Code, model checkpoints, and the EURO-5K dataset are publicly available (see Data and Code Availability).

\subsection{Cross-Dataset Evaluation}
Our goal is to assess whether task-specific fine-tuning produces genuinely specialized extractors or merely general regulatory classifiers. We evaluate on two independent external corpora that probe complementary axes of generalization beyond our curated EURO-5K dataset: \emph{specificity} (correctly rejecting general regulatory statements that are not reporting obligations) and \emph{sensitivity} (recall on confirmed reporting obligations in an out-of-distribution domain).

For specificity, we use an independent EU legislative corpus~\citep{Brandsma2025ComputationalIOB}. It consists of 7,200 manually labeled sentences stratified across 52 adoption years (1971-2022) and 20 policy areas. Binary labels distinguish general regulatory statements (containing IGT's Attribute, Deontic, and Aim) from non-regulatory text. We evaluate on their held-out test set (approximately 1,450 sentences).

For sensitivity, we use the financial regulatory corpus of \citet{chuor2025}, a publicly released annotated dataset of 257 confirmed reporting obligations. As a high-vertical, out-of-distribution domain (finance) annotated independently of EURO-5K, it tests whether our models recall reporting obligations that they have never seen during training.

We notice a task mismatch, our models are trained on specialized reporting obligations and the corpus covers broader regulatory statements including compliance obligations, permissions, and prohibitions. As such, we anticipate performance degradation as models may flag general regulatory statements as reporting obligations (false positives) or miss obligations phrased differently than EURO-5K examples (false negatives).
We aim to: (1) test whether models extract reporting obligations specifically rather than regulatory statements generally, (2) assess performance under distribution shift, (3) evaluate cross-dataset consistency of training strategies, and (4) assess zero-shot recall in a specialized out-of-distribution domain.

\section{Results and Discussion} \label{sec:results}

We present comparative performance of discriminative token classification and generative fine-tuning on reporting obligation extraction, then discuss implications in context of prior work and sustainability considerations. All results are computed on the held-out test set.

\subsection{Model Performance}

Table~\ref{tab:all_results} presents performance across all approaches on the EURO-5K held-out test set (788 examples: 263 reporting obligations, 525 negatives). 

\begin{table}[h!]
\centering
\caption{Performance on EURO-5K test set. \textit{RI} indicates relative improvement in F1 points over RegEx baseline. Bold F1 scores highlight results with error bars not overlapping baseline methods. BERT: mean$\pm$std across 5 seeds; LLMs: mean$\pm$bootstrap std (n=1,000 iterations).}
\label{tab:all_results}
\small
\setlength{\tabcolsep}{5pt} 
\begin{tabular}{@{}lcccc@{}}
\toprule    
\textbf{Method} & \textbf{Precision} & \textbf{Recall} & \textbf{F1} & \textbf{RI} \\
\midrule
\multicolumn{5}{l}{\textbf{Baselines}} \\
\quad RegEx/Keyword & 0.909 & 0.228 & 0.365 & -- \\
\quad Dependency Parsing & 0.710 & 0.745 & 0.727 & +0.362 \\
\quad Few-Shot Prompting & 0.642 & 0.935 & 0.762 & +0.397 \\
\midrule
\multicolumn{5}{l}{\textbf{Discriminative (Token Classification)}} \\
\quad BERT-base (FFT)   & 0.818 & 0.920 & \textbf{0.865} $\pm$ 0.024 & +0.500 \\
\quad BERT-base (LoRA)  & 0.667 & 0.920 & 0.773 $\pm$ 0.012 & +0.408 \\
\quad Legal-BERT (FFT)  & 0.850 & 0.921 & \textbf{0.883} $\pm$ 0.022 & +0.518 \\
\quad Legal-BERT (LoRA) & 0.695 & 0.921 & 0.791 $\pm$ 0.043 & +0.426 \\
\midrule
\multicolumn{5}{l}{\textbf{Generative (Span Extraction)}} \\
\quad Llama-3.1-8B (QLoRA) & 0.867 & 0.916 & \textbf{0.891} $\pm$ 0.019 & +0.526 \\
\quad Mistral-7B (QLoRA)   & 0.858 & 0.875 & \textbf{0.866} $\pm$ 0.019 & +0.501 \\
\quad Saul-7B (QLoRA)      & 0.833 & 0.909 & \textbf{0.869} $\pm$ 0.020 & +0.504 \\
\bottomrule
\end{tabular}
\end{table}

Relative improvement (RI) quantifies F1 gains over the RegEx baseline:

\begin{equation}
RI = F_1^{\text{method}} - F_1^{\text{RegEx}}
\end{equation}

Simple pattern matching achieves limited recall despite high precision, showing that keyword patterns identify obligations reliably but miss contextual variations. Dependency parsing greatly improves performance (+36 points) through grammatical structure analysis. Few-shot prompting performs comparably to dependency parsing, demonstrating that pre-trained language models can leverage in-context learning for specialized tasks.

Supervised fine-tuning provides considerable gains. Generative models fine-tuned with QLoRA perform comparably to discriminative FFT: Llama surpasses both BERT variants. Among 7B models, Mistral and legally-tuned Saul achieve similar performance, demonstrating minimal impact of legal adaptation at this scale. Generic Llama-8B outperforms all 7B variants, showing model scale provides benefits beyond domain adaptation. BERT-base LoRA shows the accuracy-efficiency trade-off, achieving lower F1 than FFT while using much fewer parameters.

Llama achieves the highest F1, demonstrating that generative QLoRA can surpass discriminative FFT for specialized extraction. Legal-BERT FFT closely follows, with the gap falling within statistical uncertainty. All neural approaches clearly outperform baselines, with non-overlapping error bars confirming clear performance separation.

Table~\ref{tab:confusion_matrices} presents confusion matrices revealing error trade-offs. RegEx achieves exceptional precision but misses most positives (high FN), while Few-Shot maximizes recall at the cost of many false positives. Fine-tuned models achieve better balance, with Legal-BERT LoRA showing highest sensitivity. All neural approaches substantially reduce false negatives compared to RegEx, demonstrating effective obligation detection.

\begin{table*}[ht!]
\centering
\begin{tabular}{@{}lrrrr@{}}
\toprule
\textbf{Model} & \textbf{TP} & \textbf{FP} & \textbf{FN} & \textbf{TN} \\
\midrule
\multicolumn{5}{l}{\textit{Baselines}} \\
RegEx                     &  60 &   6 & 203 & 519 \\
Dependency                & 196 &  80 &  67 & 445 \\
Few-Shot                  & 246 & 137 &  17 & 388 \\
\midrule
\multicolumn{5}{l}{\textit{Discriminative Models}} \\
BERT-base (FFT)           &   244 &    24 &    19 &   501 \\
BERT-base (LoRA)          &   248 &    34 &    15 &   491 \\
Legal-BERT (FFT)          &   247 &    28 &    16 &   497 \\
Legal-BERT (LoRA)         &   251 &    39 &    12 &   486 \\
\midrule
\multicolumn{5}{l}{\textit{Generative Models}} \\
Llama-8B (QLoRA)          & 241 &  37 &  22 & 488 \\
Mistral-7B (QLoRA)        & 230 &  38 &  33 & 487 \\
Saul-7B (QLoRA)           & 239 &  48 &  24 & 477 \\
\bottomrule
\end{tabular}
\caption{Confusion matrices for all approaches on EURO-5K test set (n=788). BERT values show mean across 5 seeds.}
\label{tab:confusion_matrices}
\end{table*}

\subsection{Comparative Analysis}

\textbf{Discriminative vs Generative Trade-offs.} Generative QLoRA (Llama) slightly surpasses discriminative FFT (Legal-BERT), reversing the traditional advantage. With proper optimization, extraction via generation matches or exceeds token classification.

\noindent\textbf{Parameter Efficiency vs Semantic Understanding.} Llama-QLoRA substantially outperforms BERT-base LoRA (+12 points) despite similar trainable parameters ($\approx$3-4\%), revealing that foundation model capacity outweighs adapter size. This demonstrates that pretraining quality and model scale matter more than parameter count for specialized legal extraction.

\noindent\textbf{Fine-tuning vs Prompting.} Fine-tuned Llama QLoRA achieves 13-point improvement over few-shot prompting, demonstrating that task-specific fine-tuning provides clear gains even over strong prompting baselines.

\subsection{Statistical Significance}

To assess statistical significance of performance differences, we conducted Welch's t-test for multi-seed BERT comparisons (accounting for unequal variances) and bootstrap paired tests (10,000 iterations) for cross-model comparisons. Table~\ref{tab:significance} presents results for theoretically motivated comparisons.

\begin{table}[h]
\centering
\small
\caption{Statistical significance of key model comparisons. Diff indicates mean F1 difference.}
\label{tab:significance}
\setlength{\tabcolsep}{6pt}
\begin{tabular}{@{}lccc@{}}
\toprule
\textbf{Comparison} & \textbf{Method} & \textbf{Diff} & \textbf{p-value} \\
\midrule
\multicolumn{4}{l}{\textit{Domain Adaptation (Welch's t-test, n=5 seeds)}} \\
\quad Legal-BERT FFT vs BERT-base FFT   & Welch-t & +0.018 & 0.307 \\
\quad Legal-BERT LoRA vs BERT-base LoRA & Welch-t & +0.018 & 0.454 \\
\midrule
\multicolumn{4}{l}{\textit{Training Strategy (Welch's t-test, n=5 seeds)}} \\
\quad BERT-base: FFT vs LoRA        & Welch-t & +0.092 & \textbf{<0.001$^{\ddag}$} \\
\quad Legal-BERT: FFT vs LoRA       & Welch-t & +0.092 & \textbf{0.009$^{\dag}$} \\
\midrule
\multicolumn{4}{l}{\textit{Paradigm Comparison (Bootstrap, n=10K)}} \\
\quad Llama vs Legal-BERT FFT       & Bootstrap & +0.008 & 0.082 \\
\quad Mistral vs Legal-BERT FFT     & Bootstrap & --0.052 & \textbf{0.001$^{\ddag}$} \\
\quad Saul vs Legal-BERT FFT        & Bootstrap & --0.049 & \textbf{<0.001$^{\ddag}$} \\
\bottomrule
\multicolumn{4}{l}{\footnotesize * $p<0.05$, $^{\dag} p<0.01$, $^{\ddag} p<0.001$} \\
\end{tabular}
\end{table}

Results confirm that FFT significantly outperforms LoRA across both base models (p<0.01), validating the accuracy-efficiency trade-off. Critically, domain adaptation effects (1.8-point gaps) lack statistical significance (p>0.30), indicating that systematic hyperparameter optimization enables generic models to approach domain-adapted performance. Llama achieves statistical parity with BERT-FFT models (p=0.08), confirming near-equivalence between paradigms for specialized extraction. Mistral and Saul perform significantly worse than Llama (p<0.01), demonstrating scale effects beyond domain adaptation.

\subsection{Inter-Model Agreement}

To assess whether paradigms learn complementary patterns, we calculate Krippendorff's $\alpha$ across model predictions. Within-paradigm agreement is high ($\alpha$=0.886 for BERT variants), indicating consistent learning across training strategies. Cross-paradigm agreement is moderate ($\alpha$=0.381), suggesting discriminative and generative approaches extract partially overlapping but distinct obligation sets. This complementarity aligns with findings from prior legal NLP work~\citep{Brandsma2025ComputationalIOB}, where cross-paradigm $\alpha$=0.58 for broader regulatory detection. The lower agreement in our specialized task reflects that paradigms emphasize different linguistic cues, as shown in our explainability analysis.

\subsection{Data Efficiency Analysis}

\begin{figure}[h]
\centering
\includegraphics[width=0.95\columnwidth]{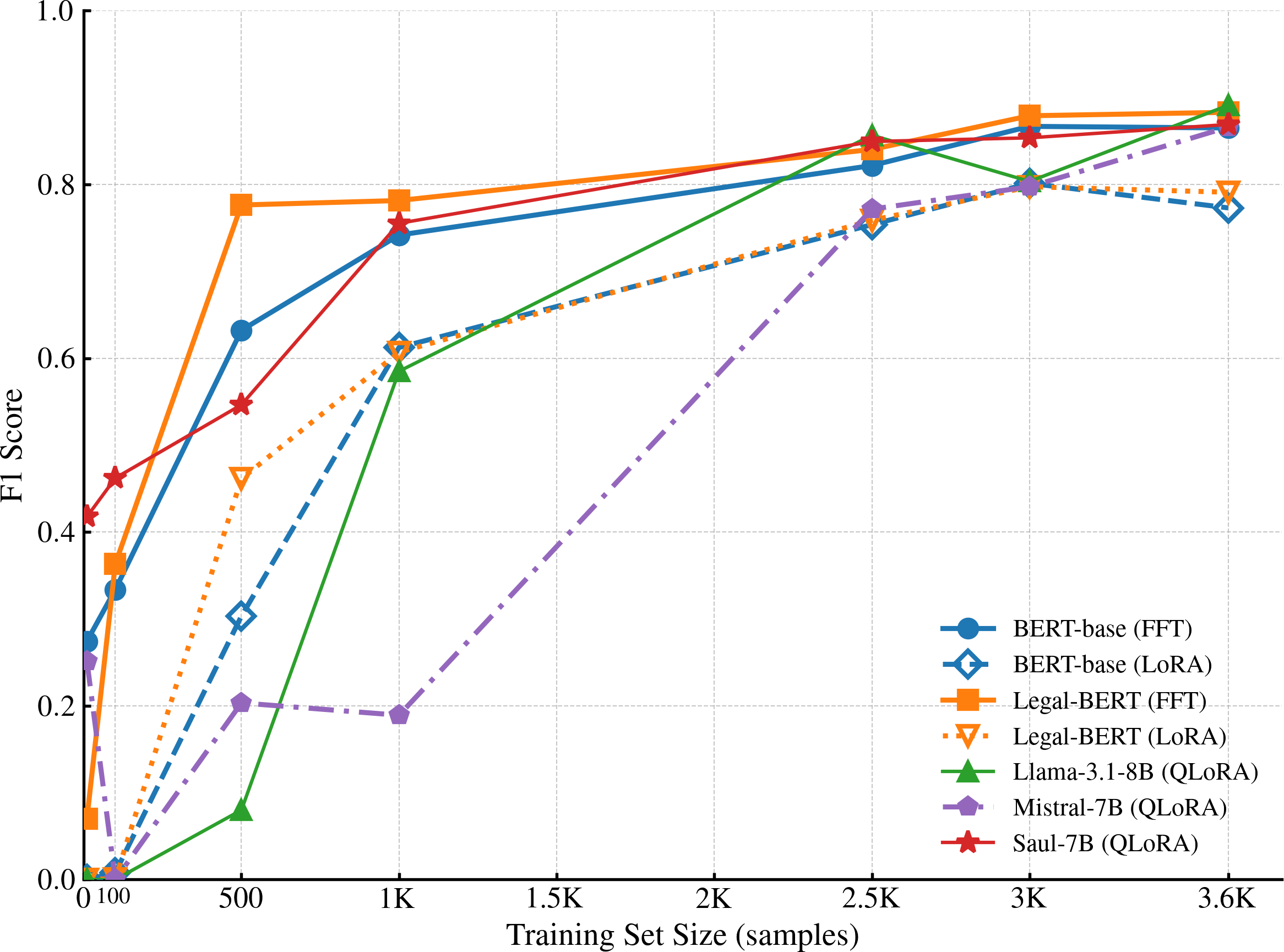}
\caption{Learning curves showing F1 performance versus training set size. Models exhibit distinct minimum data requirements and convergence patterns. Legal pretraining (Saul) reduces data needs for small datasets.}
\label{fig:learning_curves}
\end{figure}

To assess minimum data requirements, we evaluate learning curves across variable training set sizes for representative models from each paradigm and training strategy. Figure~\ref{fig:learning_curves} shows performance scaling with data availability.

Learning curves reveal distinct minimum data thresholds across training strategies. Full fine-tuning models achieve non-zero performance from 100 samples, while parameter-efficient methods require substantially more data to begin learning. Among generative models, Llama needs over 1K samples for meaningful performance, while Saul demonstrates earlier competence with minimal data.

Legal pretraining shows clearest benefit with extreme data scarcity. Saul achieves nearly half of its full performance with only 10 samples, while generic models struggle significantly at this scale. This advantage diminishes as data increases: beyond 2K samples, all models converge to similar performance levels, suggesting domain pretraining primarily accelerates early learning rather than improving final performance.

All models show substantially reduced learning rates beyond 3K samples, with gains of 3-5 points to full dataset (0.5-0.8 points per 100 samples) compared to steeper gains below 3K. This reduced rate suggests our dataset size is sufficient for competitive extraction performance, indicating diminishing returns from further data collection. Minor discrepancies between learning curve points and full dataset results reflect training variance rather than systematic trends.

\subsection{Domain Adaptation Analysis}

We compare base models against legally pretrained variants to assess domain pre-training value. Results reveal that with optimal hyperparameters, domain adaptation provides consistent but modest benefits regardless of training strategy.

\textbf{Discriminative: FFT.} Legal-BERT surpasses BERT-base by 1.8 points (88.3\% vs 86.5\%), though this difference lacks statistical significance (p=0.31, Welch's t-test). With optimized training, legal pretraining provides modest but not significant benefit. The narrow gap demonstrates that systematic hyperparameter optimization enables generic BERT to learn task patterns effectively, reducing dependence on domain-specific initialization.

\textbf{Discriminative: LoRA.} The gap remains consistent at 1.8 points (Legal-BERT: 79.1\%, BERT-base: 77.3\%, p=0.45). Remarkably, optimization produces identical gaps for both FFT and LoRA, offering no evidence that parameter constraints amplify pretraining dependence. This suggests proper hyperparameter selection matters more than parameter budget. Higher Legal-BERT variance (±4.3\% vs ±1.2\%) indicates less stable optimization despite domain initialization.

\textbf{Generative: QLoRA.} Legal-tuned Saul achieves negligible benefit over base Mistral (0.3-point gap, within statistical noise). At 7B scale with consistent hyperparameters, legal adaptation provides minimal benefit. Generic Llama-8B clearly outperforms both 7B variants (+2-3 points), indicating model scale outweighs domain adaptation. However, Llama's superiority conflates scale effects (12.5\% more parameters) with architectural and pretraining differences. Isolating pure domain adaptation effects at LLM scale would require matched-scale comparison (e.g., Legal-Llama-8B vs base Llama-8B); our matched-scale evidence comes from the 7B comparison (Saul vs Mistral).

\subsection{Cross-Dataset Evaluation}

To validate task specialization, we evaluate models on an external EU regulatory corpus~\citep{Brandsma2025ComputationalIOB}. Table~\ref{tab:crossdata} compares performance across datasets.

\begin{table}[h]
\centering
\small
\caption{Cross-dataset evaluation on external regulatory corpus (n=1,451). \textit{Pred+}: Total positive predictions. \textit{True Prec.}: Manual validation of 25 predictions per model confirms actual reporting obligations (145/175 valid overall, 82.9\%).}
\label{tab:crossdata}
\setlength{\tabcolsep}{4pt} 
\begin{tabular}{@{} l S[table-format=1.3] S[table-format=1.3] S[table-format=1.3] c S[table-format=1.3] @{}}
\toprule
\textbf{Model} & {\textbf{Precision}} & {\textbf{Recall}} & {\textbf{F1}} & \textbf{Pred+} & {\textbf{True Prec.}} \\
\midrule
BERT-base (FFT)      & 0.888 & 0.145 & 0.250 & 116 & 0.920 \\
BERT-base (LoRA)     & 0.890 & 0.171 & 0.287 & 136 & 0.800 \\
Legal-BERT (FFT)     & 0.868 & 0.158 & 0.268 & 129 & 0.760 \\
Legal-BERT (LoRA)    & 0.923 & 0.169 & 0.286 & 130 & 0.760 \\
\midrule
Llama-3.1-8B (QLoRA) & 0.966 & 0.119 & 0.211 & 87  & 0.920 \\
Mistral-7B (QLoRA)   & 0.938 & 0.150 & 0.258 & 113 & 0.880 \\
Saul-7B (QLoRA)      & 0.886 & 0.165 & 0.279 & 132 & 0.760 \\
\bottomrule
\end{tabular}
\end{table}

Manual validation of 175 predictions (25 per model) reveals that 82.9\% (145/175) are actual reporting obligations, confirming task-specific learning. The 30 remaining predictions identify related regulatory constructs rather than random errors: 40\% flag behavioral obligations (operational requirements without reporting), 25\% capture disclosure requirements (public communication rather than authority reporting), 15\% mark content specifications, with the remainder involving procedural coordination or permission statements. Critically, zero predictions were completely non-regulatory, demonstrating that models learned regulatory language patterns while maintaining reporting obligation focus. Models appropriately achieve low recall (12-17\%), correctly rejecting the behavioral and disclosure obligations that constitute the majority of the external corpus.

The performance gap between datasets (F1: 0.87 in-domain, 0.25 cross-domain) reflects task specialization rather than poor generalization. Models extract reporting obligations specifically entity-to-authority submissions for supervisory purposes. At the same time, they correctly reject broader regulatory statements in the external corpus that lack this oversight dimension. This validates our specialized approach that models behave as reporting obligation extractors, not generic regulatory classifiers.

True precision varies across models, with generic models (BERT-base FFT: 92\%, Llama: 92\%) showing higher specificity than legal-pretrained variants (Legal-BERT: 76\%, Saul: 76\%). Wilson score confidence intervals (95\%) for individual models range from [75-98\%] for high-precision models to [57-89\%] for moderate-precision models, providing adequate statistical power to distinguish behavioral patterns despite the 25-sample validation size. This pattern suggests domain pretraining may increase sensitivity to related regulatory constructs, capturing a wider range of obligation like statements while maintaining strong focus on reporting requirements.

The specificity result above shows that our models correctly reject regulatory statements that are not reporting obligations. To complete the picture, we assess \emph{sensitivity}: whether the same models recall \emph{confirmed} reporting obligations in an out-of-distribution domain. We evaluate zero-shot on the 257 financial reporting obligations of \citet{chuor2025}, a corpus annotated independently of EURO-5K. Table~\ref{tab:chuor} reports per-model recall.

\begin{table}[h]
\centering
\small
\caption{Zero-shot recall on the external financial reporting corpus of \citet{chuor2025} ($n=257$ confirmed reporting obligations). As the corpus has only positives, recall is the meaningful metric (Precision is trivially 1.000 and F1 inflated).}
\label{tab:chuor}
\setlength{\tabcolsep}{8pt}
\begin{tabular}{@{} l S[table-format=1.3] S[table-format=1.3] S[table-format=3.0] S[table-format=2.0] @{}}
\toprule
\textbf{Model} & {\textbf{Recall}} & {\textbf{F1}} & {\textbf{Detected}} & {\textbf{Missed}} \\
\midrule
BERT-base (FFT)      & 0.840 & 0.913 & 216 & 41 \\
BERT-base (LoRA)     & 0.887 & 0.940 & 228 & 29 \\
Legal-BERT (FFT)     & 0.864 & 0.927 & 222 & 35 \\
Legal-BERT (LoRA)    & 0.875 & 0.934 & 225 & 32 \\
\midrule
Llama-3.1-8B (QLoRA) & 0.677 & 0.807 & 174 & 83 \\
Mistral-7B (QLoRA)   & 0.879 & 0.936 & 226 & 31 \\
Saul-7B (QLoRA)      & 0.903 & 0.949 & 232 & 25 \\
\bottomrule
\end{tabular}
\end{table}

Despite no exposure to financial regulation during training, the models recall the majority of obligations, reaching 90.3\% (Saul-7B) and 88.7\% (BERT-base LoRA). Notably, parameter-efficient adaptation transfers better than full fine-tuning: BERT-base (LoRA) improves over its FFT counterpart (0.887 vs.\ 0.840), consistent with LoRA preserving more of the base model's general language competence and overfitting less to the EURO-5K writing style. The discriminative (BERT) figures are deterministic, whereas the generative (LLM) figures reflect a single decoding run and may vary slightly under stochastic generation. Qualitative inspection of the missed cases mirrors the previous error analysis. Of the 29 cases missed by BERT-base (LoRA), most are constructs our taxonomy deliberately excludes: inter-institutional notifications (\emph{``The Commission shall inform the Member States of the measures taken''}), legislative proposals (\emph{``the Commission shall submit to the European Parliament and to the Council any additional proposal\dots''}), and public reports (\emph{``the Commission shall publish a general report on the experience acquired\dots''}), which under our stricter definition (Definition~\ref{def:ro}) fall outside entity-to-authority supervisory reporting. A minority are genuine reporting obligations phrased atypically as applications or permissive notifications (\emph{``the marketing authorisation holder may notify the Agency\dots''}), indicating residual room for recall gains on weakly-deontic formulations. Effective recall on in-scope obligations is therefore higher than the raw figures suggest. Together with the specificity result, this confirms that EURO-5K fine-tuning captures the semantic structure of reporting obligations across legal domains, rather than fitting the training distribution.

\subsection{Explainability Analysis}

To understand model decision making, we analyze feature importance using LIME~\citep{Ribeiro2016LIME,Dianna2022} for BERT models and attention weights for LLMs on 50 balanced test examples (25 positive, 25 negative). For aggregate statistics and visualizations, we filter common words (e.g., 'the', 'and', 'of'). Table~\ref {tab:feature_importance} presents the top 10 features by cumulative importance across all models.

\begin{table}[h]
\centering
\caption{Top 10 features ranked by aggregate importance across models (sum of percentages). Percentages show frequency in top-ranked features across TP examples.}
\label{tab:feature_importance}
\begin{tabular}{@{} l cc cc ccc @{}}
\toprule
& \multicolumn{2}{c}{\textbf{Legal-BERT}} & \multicolumn{2}{c}{\textbf{BERT-base}} & \multicolumn{3}{c}{\textbf{LLMs}} \\
\cmidrule(lr){2-3} \cmidrule(lr){4-5} \cmidrule(lr){6-8}
Feature & FFT & LoRA & FFT & LoRA & Llama & Mistral & Saul \\
\midrule
Commission & 37.5 & 36.0 & 39.1 & 40.0 & 16.7 & 21.7 & 13.0 \\
Member & 12.5 & 16.0 & 13.0 & 16.0 & 29.2 & 39.1 & 30.4 \\
Shall & 8.3 & 8.0 & 13.0 & 8.0 & 37.5 & 30.4 & 26.1 \\
Authority & 16.7 & 16.0 & 17.4 & 16.0 & 16.7 & 13.0 & 21.7 \\
Regulation & 12.5 & 8.0 & 8.7 & 12.0 & 16.7 & 17.4 & 26.1 \\
Annex & 16.7 & 16.0 & 17.4 & 12.0 & 8.3 & 13.0 & 13.0 \\
Authorities & 8.3 & 8.0 & --- & 8.0 & 12.5 & 13.0 & 13.0 \\
Report & 8.3 & 8.0 & 13.0 & 12.0 & 12.5 & 8.7 & --- \\
Competent & 12.5 & 12.0 & 13.0 & 12.0 & --- & --- & 8.7 \\
Out & 12.5 & 16.0 & 17.4 & 12.0 & --- & --- & --- \\
\bottomrule
\end{tabular}
\end{table}

All models consistently emphasize institutional actors and regulatory frameworks. \textit{Commission} appears as a top feature across BERT models and also in LLMs, reflecting its central role in EU reporting. \textit{Authority} shows cross-architecture consensus, validating that both paradigms learn core regulatory entities.

Despite consensus on institutional terms, BERT and LLM models exhibit distinct emphasis patterns. LLMs demonstrate much stronger focus on deontic markers and entity references, e.g., \textit{Shall}, \textit{Member}, \textit{Regulation}. This architectural difference may reflect LLMs' causal language modeling objective, which emphasizes action sequences and procedural flow.

Terms like \textit{shall} and \textit{Member State} also appear with comparable frequency in both correct identifications and correct rejections, but with opposite contribution scores in LIME analysis. This pattern indicates models evaluate semantic context rather than relying on surface keywords. 

\begin{figure}[h!]
\centering
\includegraphics[width=0.95\textwidth]{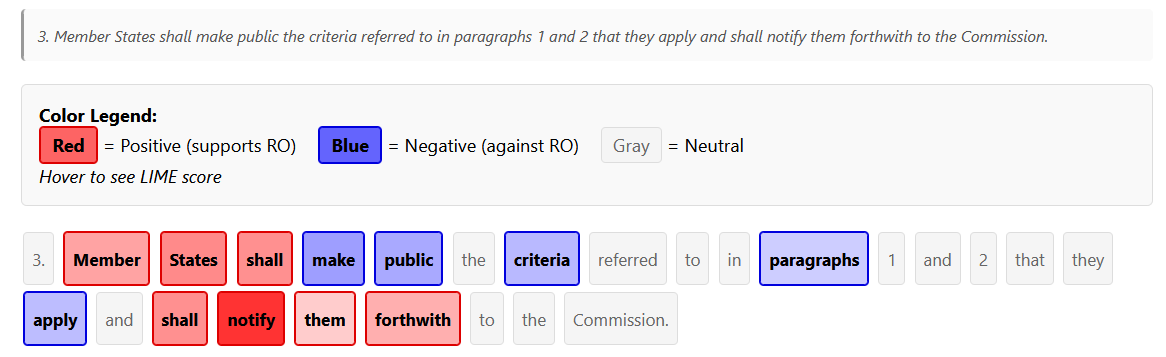}
\caption{Explainability visualizations (LIME) for a reporting obligation in Par. 3, Art. 64 of Reg. (EU) 1308/2013. The model assigns positive weights (red) to the transmission action (\textit{notify}), while assigning negative weights (blue) to disclosure actions (\textit{public}), correctly resolving intra-sentence functional ambiguity.}
\label{fig:explainability}
\end{figure}

Figure~\ref{fig:explainability} illustrates this contextual evaluation through a correctly identified reporting obligation, containing multiple \textit{shall} clauses, from paragraph 3 of Article 62 of Regulation (EU) No 1308/2013 (CELEX:32013R1308). The visualization shows strong positive contributions for \textit{notify}, while \textit{public} receives negative weight, as publication constitutes disclosure rather than reporting. The model correctly distinguishes \textit{"shall notify... Commission"}, a reporting obligation, from \textit{"shall make public..."}, a disclosure obligation, within the same sentence, demonstrating learned contextual sensitivity beyond keyword matching.

Legal pretraining affects feature prioritization patterns. Legal-BERT models consistently assign higher importance to structural references (e.g., Annex) and procedural terms (e.g., delay) compared to Base-BERT models. However, this specialization does not uniformly translate to improved performance, as the 1.8-point F1 gap indicates.

Explainability analysis reveals that model architectures and training objectives shape feature importance patterns even when achieving similar accuracy. This suggests that model selection should consider not only performance metrics but also the interpretative requirements of the deployment context. For instance, BERT's concentrated feature attribution may better suit compliance verification, while LLMs' distributed attention patterns may better capture complex multi-clause obligations.

\subsection{Sustainability}

Parameter-efficient adaptation (LoRA for BERT, QLoRA for LLMs) trains 2.3\% of parameters versus FFT's 100\%, reducing memory requirements. Energy measurements via CodeCarbon~\citep{codecarbon2024,strubell2019energy} show BERT training consumes 0.065-0.076 kWh per 5-seed run for both FFT and LoRA, while LLM QLoRA training requires 0.19-0.21 kWh per single-seed run.

Comparable consumption between parameter-efficient and full fine-tuning reflects a time-energy trade-off: adapter computations increase training duration, offsetting parameter savings. The primary value lies in enabling memory-constrained deployment rather than energy reduction.

\section{Practical Applications} \label{sec:applications}

To demonstrate practical deployment feasibility, we developed an end-to-end system for reporting obligation extraction comprising an interactive web interface with integrated RDF export and a REST API for programmatic access.

\begin{figure}[h]
\centering
\includegraphics[width=0.95\columnwidth]{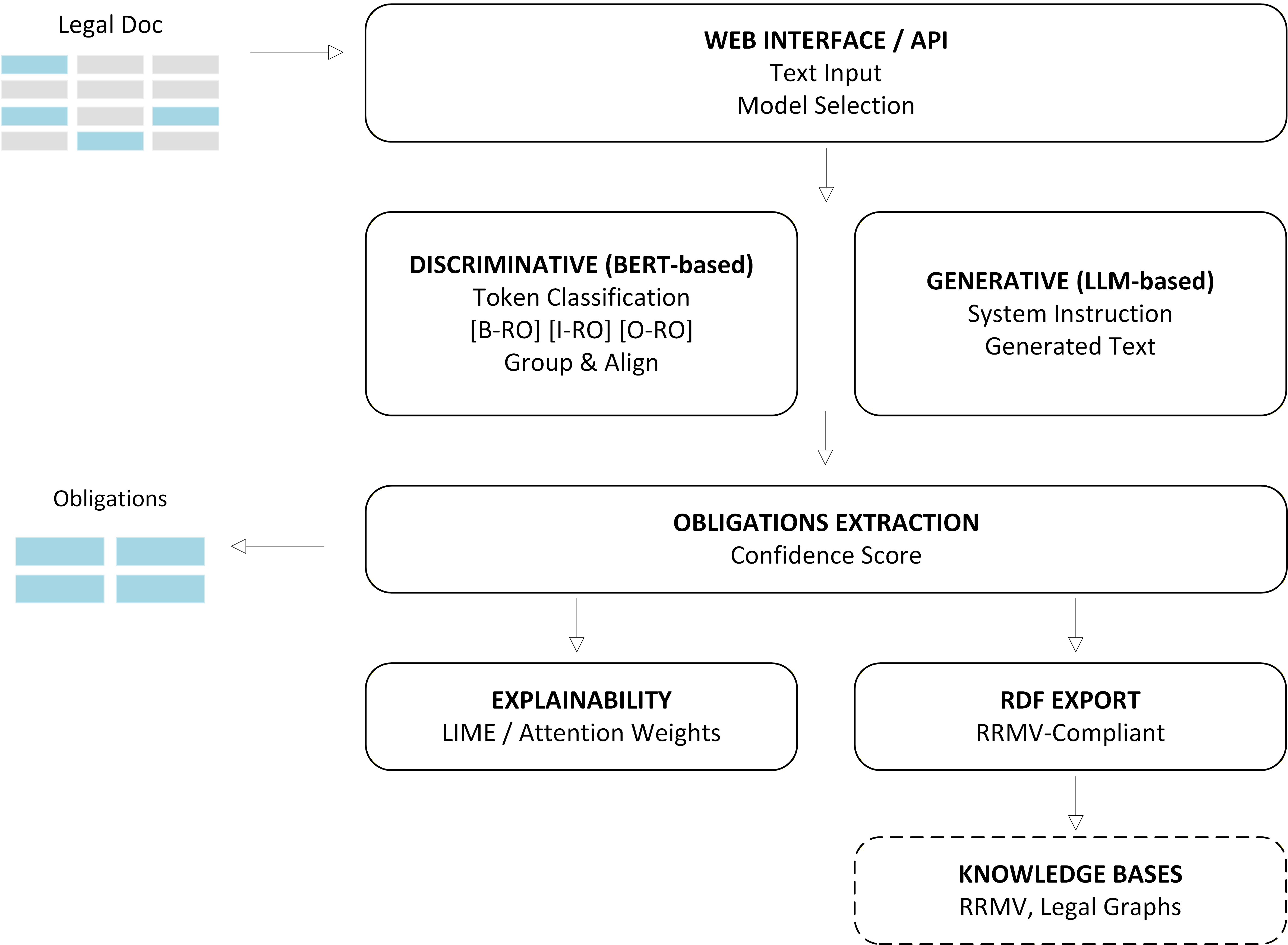}
\caption{Extraction workflow showing paradigm selection, processing paths, and downstream features (explainability and RDF export)}
\label{fig:extraction_workflow}
\end{figure}

Figure~\ref{fig:extraction_workflow} illustrates the extraction workflow. Users input legislative text and select an extraction paradigm (discriminative token classification or generative span extraction). The system processes text through the selected approach, extracts reporting obligations, and provides explainability visualizations (LIME for BERT, attention for LLMs) along with optional RDF export for regulatory knowledge bases. 

\subsection{Interactive Web Interface} 

The web interface provides exploration of model capabilities, combining extraction, explainability, and structured export in a unified workflow accessible to users without technical expertise.

\textbf{Input and Model Selection.} Users paste legislative text into a text box and select a trained model from available options (seven models in total). The interface supports two analysis modes: (1) full-text analysis processing the entire input as-is, or (2) sentence-split analysis where the system automatically segments text into sentences and processes each independently. This flexibility accommodates both targeted single-sentence analysis and multi-paragraph document processing.

\textbf{Extraction Results.} For positive classifications, the interface highlights extracted obligation spans within the input text using color coding. Confidence scores accompany predictions, derived from model output probabilities for BERT (softmax over BIO tags) and generation likelihood for LLMs (token probability sequences). These scores help users assess prediction reliability and prioritize manual review for low-confidence cases. Users can compare predictions across models to assess consensus and identify uncertain cases requiring expert validation.

\textbf{Explainability Visualizations.} The interface integrates interpretability tools to support result understanding. For BERT models, LIME generates feature importance scores showing which terms contributed positively (highlighted in red) or negatively (highlighted in blue) to the classification decision. For LLMs, attention weight visualizations use gradient coloring to indicate token relevance during generation. While LIME and attention capture different explanatory aspects (prediction sensitivity versus internal computation flow), both help users understand textual cues driving model decisions and assess whether predictions rely on appropriate linguistic patterns.

\textbf{RDF Export.} For detected reporting obligations, the interface provides RRMV-compliant RDF export functionality. Users can generate structured representations, exporting results in RDF/Turtle format compatible with the EC's Reporting Requirements Metadata Vocabulary. The converter applies pattern-based entity recognition to identify reporting entities (who must submit) and recipient authorities (who receives information), then generates RRMV triples encoding these regulatory relationships. This enables immediate integration with knowledge bases and compliance management systems, demonstrating end-to-end workflow from extraction to structured knowledge representation.

\begin{figure}[ht]
\centering
\includegraphics[width=0.95\columnwidth]{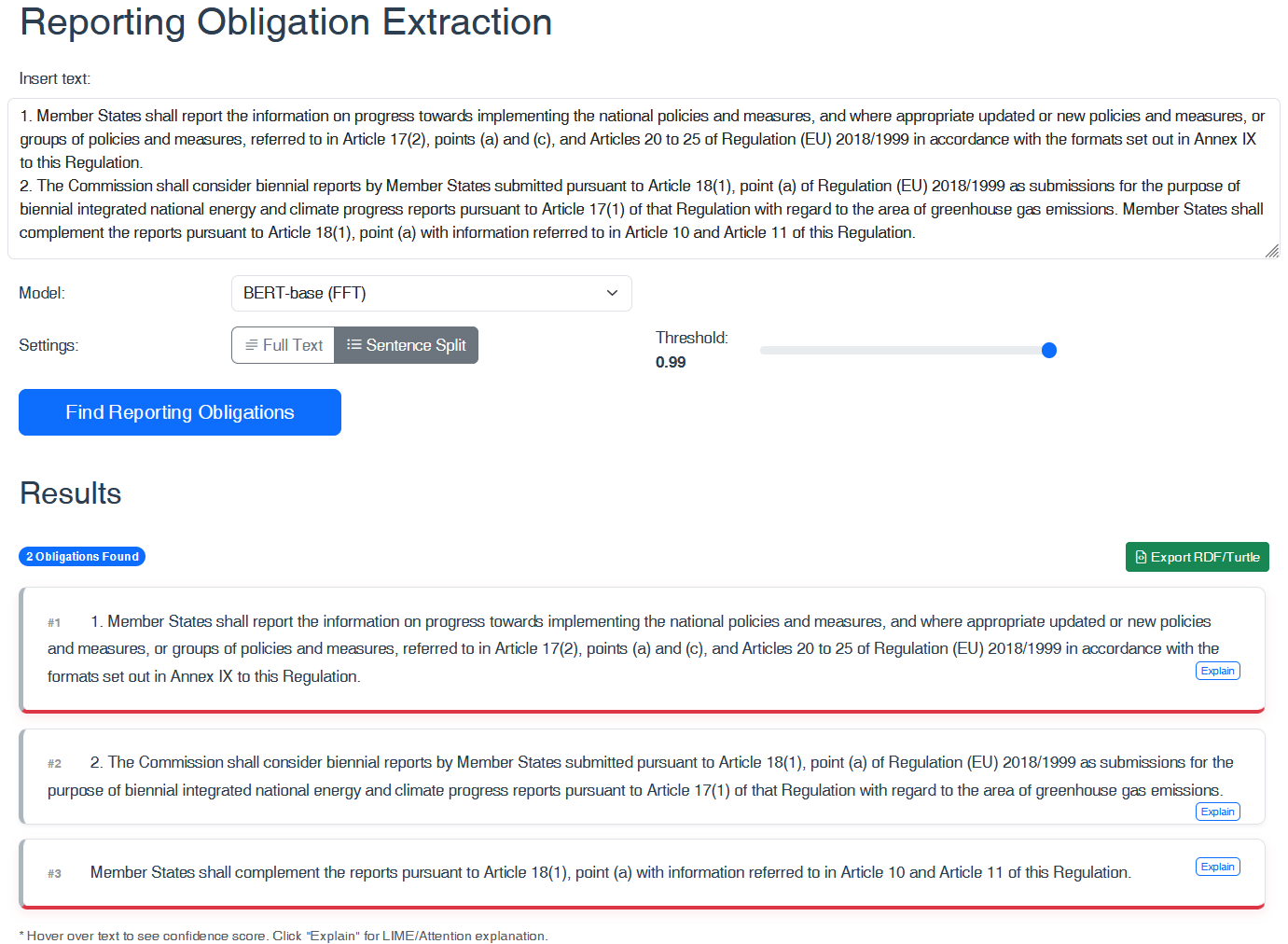}
\caption{Web-based interface for interactive reporting obligation extraction. Features include model selection, dual analysis modes, confidence scoring, explainability visualizations, and RRMV-compliant RDF export.}
\label{fig:demo_ui}
\end{figure}

Figure~\ref{fig:demo_ui} shows the interface design integrating these capabilities. The prototype is designed as a research artifact demonstrating deployment feasibility. Source code and live demo will be made publicly available following the review period.

\subsection{REST API}

For integration with existing regulatory compliance systems, we provide a REST API enabling programmatic access to the trained models. The API architecture follows a modular design, decoupling text analysis, explainability, and structured data export.

The system exposes four primary endpoints:
\begin{enumerate}[label=(\roman*), itemsep=0.3em, topsep=0pt]
    \item \texttt{GET /api/models} for retrieving the list of supported models; 
    \item \texttt{POST /api/predict} for core extraction using configurable confidence thresholds; 
    \item \texttt{POST /api/explain} for retrieving local interpretability data (LIME or attention weights); 
    \item \texttt{POST /api/export/rdf} for transforming predictions into structured formats.    
\end{enumerate}

The prediction endpoint supports automatic sentence segmentation via the \texttt{shouldSplitSent} parameter, enabling the processing of raw legislative text. Example request structure:

\begin{small}
\begin{verbatim}
POST /api/predict
{
  "text": "Member States shall report the information...",
  "model": "bert_base_fft",
  "threshold": 0.5,
  "shouldSplitSent": true
}
\end{verbatim}
\end{small}

Response provides a structured list of predictions, including the sentence index, the extracted string, a binary classification and the model's confidence score. For downstream semantic web applications, the export endpoint transforms these results into RRMV-compliant RDF, allowing automated population of regulatory knowledge bases. This ensures interoperability with Linked Open Data (LOD) standards in the legal domain, facilitating the integration of extracted obligations into regulatory monitoring systems.

We package the system as a Docker container enabling single-command deployment. Users can select specific models to load, minimizing memory footprint for resource-constrained environments. While training requires GPUs, inference runs efficiently on CPU infrastructure, making the system accessible for public administration use cases with limited computational resources.

\subsection{RRMV-Compliant Output Format}

The RDF export functionality generates output following the Reporting Requirements Metadata Vocabulary (RRMV) specification, an EC initiative supporting the 25\% regulatory burden reduction target~\citep{rrmv}. RRMV provides a standardized ontology for cataloging reporting obligations across EU legislative acts. Our converter structures detected obligations through three steps:

\begin{enumerate}[leftmargin=*, itemsep=1em]
    \item \textbf{Obligation Detection:} Models classify sentences, filtering non-obligations and identifying reporting requirements.
    \item \textbf{Entity Recognition:} Pattern-based extraction identifies reporting entities and recipient authorities using predefined patterns for common EU regulatory actors (Commission, ESMA, EBA, ECB, Member States, competent authorities). The patterns cover both explicit mentions (``Member States shall submit to the Commission'') and common regulatory phrasings.
    \item \textbf{RDF Generation:} Extracted entities populate RRMV-compliant RDF triples encoding reporting relationships with appropriate semantic annotations.
\end{enumerate}

Consider for example the obligation: \textit{``The competent authorities of the home Member State shall provide the competent authorities of the host Member State with the organisational structure of an investment firm.''} The converter extracts the reporting entity (home Member State authorities) and recipient (host Member State authorities), then generates the following RDF:

\begin{small}
\begin{verbatim}
example:Action001 a rrmv:Action ;
  rrmv:hasAgentRole [
    rrmv:forAgent example:HomeMemberAuthorities ;
    rrmv:withRole rrmv:addresser ] ,
  [ rrmv:forAgent example:HostMemberAuthorities ;
    rrmv:withRole rrmv:addressee ] .
\end{verbatim}
\end{small}

Pattern-based entity recognition covers common EU regulatory actors but may miss novel entities or context-specific references. Complete RRMV population requires additional structured extraction of temporal specifications (deadlines, frequencies), deontic operators (obligation vs permission), content requirements, and conditional triggers. The converter addresses RRMV's primary bottleneck—finding obligations—while structured attribute extraction remains future work.

For typical EU regulatory text containing standard entities, the converter successfully generates valid RDF for 60-80\% of detected obligations. This provides substantial automation for initiatives like RRMV while acknowledging that complete regulatory catalogs require complementary manual enrichment or ML-based entity extraction.

\subsection{Deployment Workflow}

We outline a deployment workflow that integrates automated extraction with expert legal validation to support scalable regulatory analysis. Trained models are applied to legislative texts to identify candidate reporting obligations, each associated with a confidence score. Predictions exceeding a predefined confidence threshold are provisionally accepted, while lower-confidence cases are referred for expert review.

Validated obligations are automatically converted into structured representations compliant with the Reporting Requirements Metadata Vocabulary (RRMV) and integrated into regulatory knowledge bases, enabling structured querying and downstream analysis. 

This human-in-the-loop approach balances automation with legal assurance. Given the models’ high precision (85–87\%), most extracted candidates correspond to genuine reporting obligations, substantially reducing expert validation effort. Although some obligations remain undetected, the ability to automatically identify the majority represents a significant efficiency gain over manual analysis and demonstrates the practical viability of large-scale deployment.

\section{Conclusions and future work} \label{sec:conclusion}

This work evaluated generic and legal-domain transformers for reporting obligation extraction from EU legislation. Using EURO-5K, a curated dataset of 5,253 sentences, we trained seven models across discriminative and generative paradigms. All achieved high performance (F1: 0.77-0.89). Statistical testing reveals domain adaptation was not significant (p>0.05), indicating systematic optimization enables generic models to match legal-domain variants. Both paradigms achieve statistical parity. Learning curves show convergence around 3K samples while legal pretraining accelerates early learning. Cross-dataset validation confirms specialized learning, and explainability reveals emphasis on institutional actors. We demonstrate practical deployment through an interactive interface with explainability visualizations and automated RDF conversion for regulatory knowledge bases.

\subsection{Limitations}

Domain adaptation showed modest gains (1.8 F1 points) that were not statistically significant. This limits conclusions about legal pretraining benefits when systematic hyperparameter optimization is employed. LLM training used single seeds due to computational constraints. Bootstrap resampling provides confidence intervals but estimates only test set uncertainty, not training variance, likely underestimating true LLM variance. Overlapping confidence intervals between paradigms should be interpreted cautiously.

Cross-dataset validation used 25 samples per model which is adequate for pattern identification but insufficient for tight confidence bounds. The few-shot baseline used a single prompt without optimization, potentially underestimating in-context learning capabilities. Explainability analysis combined LIME and attention weights, which measure different aspects and limit direct comparison. Finally, our dataset focuses on English language EU legislation, limiting direct generalization to other legal systems. Broader jurisdictional assessment remains necessary.

\subsection{Future Work}
In future work, we plan to further study structured obligation extraction beyond binary classification to extract obligation elements (addressee, action, period, result) from detected obligations in EURO-5K. Capturing temporal conditions, deontic operators (obligation vs permission), entity roles (who reports to whom), and obligation attributes (frequency, format requirements) would enable richer compliance automation and regulatory analysis applications.

As reporting obligations function as secondary obligations tied to primary substantive duties, tracking their evolution is essential for maintaining regulatory compliance systems. When primary obligations are amended or repealed, corresponding reporting requirements must adapt accordingly. Future work should investigate automated detection of obligation modifications across legislative amendments, enabling dynamic updating of compliance monitoring systems. This capability would support Member State adherence to transparency requirements and facilitate assessment of regulatory effectiveness across the EU's evolving legislative framework.

Cross-lingual extension to other EU official languages would assess transfer learning approaches for the Union's multilingual legislative corpus. Evaluation on non-EU legislation would clarify whether our extraction methods generalize or require jurisdiction-specific adaptation. Such extensions would contribute to harmonizing compliance monitoring across Member States and regulatory domains.

Active learning approaches could reduce annotation costs for expanding the dataset. Our models achieve reasonable precision but still produce false positives requiring expert review. Uncertainty-based sampling could identify cases where models are least confident, prioritizing these for human annotation and enabling iterative dataset expansion more efficiently than random sampling.

\newpage

\begin{appendices}
\section{Representative Annotation Examples}\label{appendix_a}

This appendix presents representative annotation examples from EURO-5K curation. 
\begin{sidewaystable}[p]
\centering
\caption{Representative annotation examples from EURO-5K with decisions and key reasoning patterns.}
\label{tab:edge_cases}
\begin{tabularx}{\textheight}{@{}l>{\raggedright\arraybackslash}X c >{\raggedright\arraybackslash}X@{}}
\toprule
\textbf{ID} & \textbf{Sentence} & \textbf{Decision} & \textbf{Key Reason} \\
\midrule
1234 & ESMA shall submit those draft regulatory technical standards to the Commission by 30 September 2012. & KEEP & Authority coordination (ESMA to Commission oversight) under broader institutional definition \\
\addlinespace
636 & The annual report to be submitted by the Member States to the Commission on the compliance with sulphur standards for marine fuels shall include at least the following information: & REMOVE & Content specification: Main verb is ``include'' (content), not ``submit'' (obligation) \\
\addlinespace
3018 & All notifications to the applicant for the purpose of his or her application for a travel authorisation shall be sent to the email address provided by the applicant in the application form as referred to in point (g) of Article 17(2). & REMOVE & Wrong direction: Authority to Applicant (not upward reporting) \\
\addlinespace
7 & The competent authorities of the home Member State shall provide the competent authorities of the host Member State with the organisational structure of an investment firm, its business lines and its relationships to entities within the group. & KEEP & Peer authority coordination for cross-border supervision \\
\addlinespace
413 & The applicant shall submit to the Commission, the Member States and the Authority additional information as regards: & KEEP & Primary obligation to submit; colon introduces content enumeration \\
\addlinespace
430 & Prior to importation, the importer shall notify each consignment of the specified wood sufficiently in advance to the competent authority of the Member State of the first place of storage after arrival to the Union territory in the format laid down in Article 40(1), point (c), of Implementing Regulation (EU) 2019/1715. & KEEP & Complete reporting obligation with timing, format, and authority specification \\
\addlinespace
451 & Member States shall report the information on progress towards the objectives, targets and contributions with respect to the research, innovation and competitiveness dimension referred to in Article 4, point (e) and Article 25, points (a) to (c) of Regulation (EU) 2018/1999 in accordance with the formats set out in Annex VII to this Regulation. & KEEP & Implicit recipient (Commission contextually determined); formal reporting structure \\
\bottomrule
\end{tabularx}
\end{sidewaystable}
Table~\ref{tab:edge_cases} shows 7 examples illustrating key annotation patterns:

\begin{itemize}[leftmargin=*, nosep]
    \item \textbf{Content specifications:} Main verb specifies content not submission (Example 636)
    \item \textbf{Direction errors:} Authority to non-authority communication (Example 3018)
    \item \textbf{Institutional scope:} Authority-to-authority coordination for supervision (Examples 1234, 7)
    \item \textbf{Implicit recipients:} Contextually determined from regulatory framework (Example 451)
    \item \textbf{Obligation primacy:} Primary clause establishes duty to submit (Example 413)
\end{itemize}

\section{Dataset Curation Methodology}\label{appendix_b}

This appendix documents the semi-automated curation process transforming original AROLD annotations into EURO-5K, including the five-criteria framework, validation results, and final dataset composition.

\subsection{Overview and Pipeline}

The AROLD dataset~\citep{aroldjrc} contained 30,432 positive annotations from 136 EU regulatory documents. Initial analysis revealed systematic issues: structural noise (form labels, headers), multi-sentence annotations requiring resegmentation, and misclassifications (behavioral requirements, disclosure mandates). We implemented a multi-phase pipeline combining rule-based filtering, LLM-assisted review (Claude Sonnet 4), human validation ($\kappa$=0.613), and iterative refinement.

\subsection{Five-Criteria Framework}

A sentence qualifies as a reporting obligation if it satisfies all five criteria. Table~\ref{tab:five_criteria} presents the framework with example failure cases from EURO-5K.

\begin{sidewaystable}
\centering
\footnotesize
\caption{Five-criteria framework for reporting obligation annotation with purpose, rules, and example failure cases.}
\label{tab:five_criteria}
\newcommand{\examplefont}{\footnotesize\itshape}
\begin{tabularx}{\textheight}{
@{}
>{\hsize=0.2\hsize\centering\arraybackslash}X
>{\hsize=0.6\hsize\raggedright\arraybackslash}X
>{\hsize=0.9\hsize\raggedright\arraybackslash}X
>{\hsize=1.2\hsize\raggedright\arraybackslash}X
>{\hsize=1.9\hsize\itshape\raggedright\arraybackslash}X 
>{\hsize=1.2\hsize\raggedright\arraybackslash}X
@{}
}
\toprule
\textbf{\#} & \textbf{Criterion} & \textbf{Purpose} & \textbf{Rule} & \textbf{Example} & \textbf{Why It Fails} \\
\midrule
1 & Reporting Action & Distinguishes information submission from general obligations. & Valid sentences must contain transmission verbs (submit, report, notify, provide) directed to regulatory authorities. Main verb in primary clause determines classification. & The competent authorities referred to in Article 13 shall, upon their written request, require payment card schemes and/or payment service providers to provide all information necessary to verify the correct application of paragraphs 3 and 4 of this Article. & Main verb is ``require'' (indirect action) rather than direct transmission verb. Authority requires someone else to provide information, but does not itself report. \\
\addlinespace

2 & Mandatory & Ensures obligation is compulsory not permissive. & Valid sentences use deontic modals \textit{shall} or \textit{must}, or ``requires X to Y'' constructions. Permissive modals indicate discretion. & Where necessary for the accomplishment of the tasks of ESMA or the competent authorities in accordance with this Regulation, ESMA may, including upon the request of the competent authority of the Member States where a third country firm provides investment services or performs investment activities in accordance with this Article, ask third country firms providing services or performing activities in accordance with this Article to provide any further information in respect of their operations. & Uses permissive modal ``may'', indicating discretionary rather than mandatory action. \\
\addlinespace

3 & To Regulatory Authority & Identifies information flow direction to distinguish upward reporting from disclosure. & Valid recipients: EU agencies (Commission, ESMA), national competent authorities, peer regulatory bodies. Implicit recipients acceptable when context clear. & The competent authority of the home Member State shall inform the insurance, reinsurance or ancillary insurance intermediary in writing that the information has been received by the competent authority of the host Member State. & Information flows from authority to regulated entity (intermediary), representing downstream communication rather than upward reporting for oversight. \\
\addlinespace

4 & Information Submission & Distinguishes data/report submission from preparatory and operational activities. & Valid obligations submit substantive oversight information. Excluded: preparatory documentation, financial transactions, operational requirements. & CEIOPS shall, where necessary, provide for non-legally binding guidelines and recommendations concerning the implementation of the provisions of this Directive and its implementing measures in order to enhance the convergence of supervisory practices. & Describes operational activities (providing guidelines for supervisory convergence) rather than submitting information for oversight. \\
\addlinespace

5 & Obligation Primary & Separates primary obligation clauses from content specifications. & Main clause must establish duty to submit; content details are supplementary. & That report shall describe how and when the operator or aircraft operator has rectified or plans to rectify the non-conformities identified by the verifier and to implement recommended improvements. & Main verb (describe) specifies required content rather than establishing the obligation to submit the report. \\
\bottomrule
\end{tabularx}
\end{sidewaystable}

\subsection{LLM Prompt Structure}

The LLM review applied the five-criteria framework through a structured prompt encoding each check:
\begin{tcolorbox}[templatebox, breakable, title={LLM Curation Prompt}]

You are a legal expert analyzing EU legislation.

\textbf{CONTEXT (previous sentence):} \{context\}

\textbf{CURRENT SENTENCE:} \{text\}

\textbf{TASK:} Determine if the CURRENT sentence is a reporting obligation. 
Use context to resolve pronouns like ``They'', ``It'', ``Such entities''.

\textbf{DEFINITION:} Mandatory requirement for an entity (operator, Member State, or authority) to SUBMIT information TO a regulatory/oversight authority for supervision, enforcement, or regulatory coordination.

\textbf{CHECK ALL 5 CRITERIA:}

\textbf{1. REPORTING ACTION?}

Contains transmission verb TO/WITH authority: submit, report, notify, provide, inform, communicate, transmit, send, forward, deliver, share, supply, declare, convey, acknowledge receipt

\textit{Critical:} Focus on the MAIN VERB in the primary clause.
\begin{itemize}
    \item KEEP: ``shall make the notification'' (main verb is notification)
    \item REMOVE: ``shall use template when notifying'' (main verb is ``use'', procedural)
\end{itemize}

\textit{Note:} Both ``provide data TO authority'' and ``provide authority WITH data'' are valid.

KEEP if introduces obligation with colon (topics enumerated): ``must report on:'', ``shall notify of:'', ``shall submit:'' (main verb establishes duty, colon signals topics follow)

REMOVE if procedural/methodological requirement (HOW, not WHAT to report):
\begin{itemize}
    \item Template/format usage: ``shall use template'', ``shall follow format''
    \item Calculation methods: ``shall calculate using'', ``shall determine by''
\end{itemize}

\textbf{2. MANDATORY?}

Uses ``shall'' or ``must'' OR ``requires...to'' construction (NOT ``may'', ``can'', ``should'')

\textit{Note:} ``Convention/Regulation requires X to Y'' = mandatory obligation. Conditional obligations with ``shall/must'' still count as mandatory.

\textbf{3. TO REGULATORY AUTHORITY?}

Information submitted TO:
\begin{itemize}
    \item \textit{EU:} Commission, Council, Parliament, ESMA, EBA, ECB, EMA, EFSA, EIOPA
    \item \textit{National:} NCAs, supervisory authorities, competent authorities
    \item \textit{Peer:} Home NCA → Host NCA, Agency → Commission
\end{itemize}

\textit{Note:} Implicit recipients are valid. ``Member States shall report/submit'' typically implies reporting to Commission.

KEEP: Authority → Authority when the purpose is supervision or regulatory coordination

REMOVE: Authority → Public, Authority → private entities (applicants, consumers)

\textbf{4. INFORMATION SUBMISSION?}

About submitting actual data/reports/notifications for oversight.

REMOVE:
\begin{itemize}
    \item Preparatory documentation: ``submit monitoring plan'', ``submit methodology''
    \item Financial/operational requests: ``submit payment claims'', ``request reimbursement''
    \item Operational requirements: ``maintain records'', ``implement policy''
\end{itemize}

\textbf{5. PRIMARY PURPOSE IS OBLIGATION?}

Main clause about DUTY to submit (not content details).

KEEP: ``shall submit report including X''

REMOVE: ``Reports shall include X''

\textbf{EXAMPLES:}

\textit{KEEP (Reporting Obligations):}
\begin{enumerate}
    \item ``Member States shall submit annual reports to the Commission by March 31.''
    \item ``The competent authority shall notify ESMA of any incidents within 24 hours.''
    \item ``Member States must report on:'' (introduces obligation, topics enumerated)
\end{enumerate}

\textit{REMOVE (Not Reporting Obligations):}
\begin{enumerate}
    \item ``The Commission shall publish the findings on its website.'' (disclosure to public)
    \item ``Operators shall maintain records for five years.'' (operational/retention)
    \item ``The authority shall inform the applicant of its decision.'' (authority → private party)
\end{enumerate}

\textbf{OUTPUT:} Respond with valid JSON only. No other text.

\{\{\\
\phantom{xx}``decision'': ``KEEP'',\\
\phantom{xx}``reason'': ``All 5 checks pass - mandatory notification to Commission''\\
\}\}

or

\{\{\\
\phantom{xx}``decision'': ``REMOVE'',\\
\phantom{xx}``reason'': ``Check 2 fails - uses `may' not `shall'''\\
\}\}

\end{tcolorbox}

\subsection{Validation Results}

We validated the automated pipeline through blind dual annotation of 200 stratified samples. Table~\ref{tab:iaa_validation} presents inter-annotator agreement.

\begin{table}[h]
\centering
\caption{Inter-Annotator Agreement on 200-Sample Validation}\label{tab:iaa_validation}
\begin{tabular}{lcc}
\toprule
\textbf{Comparison} & \textbf{Agreement} & \textbf{Cohen's $\kappa$} \\
\midrule
Annotator 1 vs Annotator 2 & 80.3\% & 0.613 \\
Annotator 1 vs Automated & 81.8\% & 0.636 \\
Annotator 2 vs Automated & 90.0\% & 0.804 \\
\bottomrule
\end{tabular}
\end{table}

Inter-annotator agreement of $\kappa$=0.613 (substantial) demonstrates feasible yet complex annotation reflecting genuine interpretative challenges. The automated system achieved comparable performance ($\kappa$=0.636-0.804), validating pipeline reliability. Error analysis of unanimous disagreements (8 cases) revealed 2 false negatives and 6 false positives, indicating 96\% accuracy when annotators agreed.

\subsection{Final Dataset Composition}

The multi-phase curation yielded:
\begin{itemize}
    \item 1,751 validated single-sentence reporting obligations
    \item 3,502 negatives (including 532 hard negatives, 10.3\%)
    \item 5,253 total examples from 136 documents
    \item Class ratio: 1:2 (positives:negatives)
\end{itemize}

Document-level stratified splitting prevents data leakage. Hard negatives represent challenging cases preventing superficial pattern learning. Complete curation methodology, including LLM prompt template, is available in the code repository.

\section{Statistical Significance Testing}\label{appendix_c}

This appendix provides complete pairwise significance testing results for all seven models evaluated in this study.

\subsection{Methodology}

We assess statistical significance using two complementary approaches:

\textbf{Welch's t-test (BERT models).} For models trained with multiple random seeds (n=5), we compare F1 score distributions using Welch's t-test, which accounts for potentially unequal variances across training runs. This approach is appropriate given observed variance heterogeneity (e.g., Legal-BERT LoRA: $\sigma$=4.3\% vs BERT-base LoRA: $\sigma$=1.2\%).

\textbf{Bootstrap paired testing (Cross-model).} For comparisons involving single-seed LLM models or across paradigms, we employ bootstrap resampling with 10,000 iterations. Each iteration resamples the test set with replacement, calculates F1 scores for both models, and computes the difference. The two-tailed p-value represents the proportion of bootstrap samples where the performance difference has opposite sign to the observed difference.

All comparisons use aligned predictions on the identical 788-example test set, ensuring valid paired testing. P-values are interpreted using conventional thresholds: p<0.05 (*), p<0.01 (†), p<0.001 (‡).

\subsection*{Complete Pairwise Significance Matrix}

\begin{figure}[h!]
\centering
\includegraphics[width=0.85\textwidth]{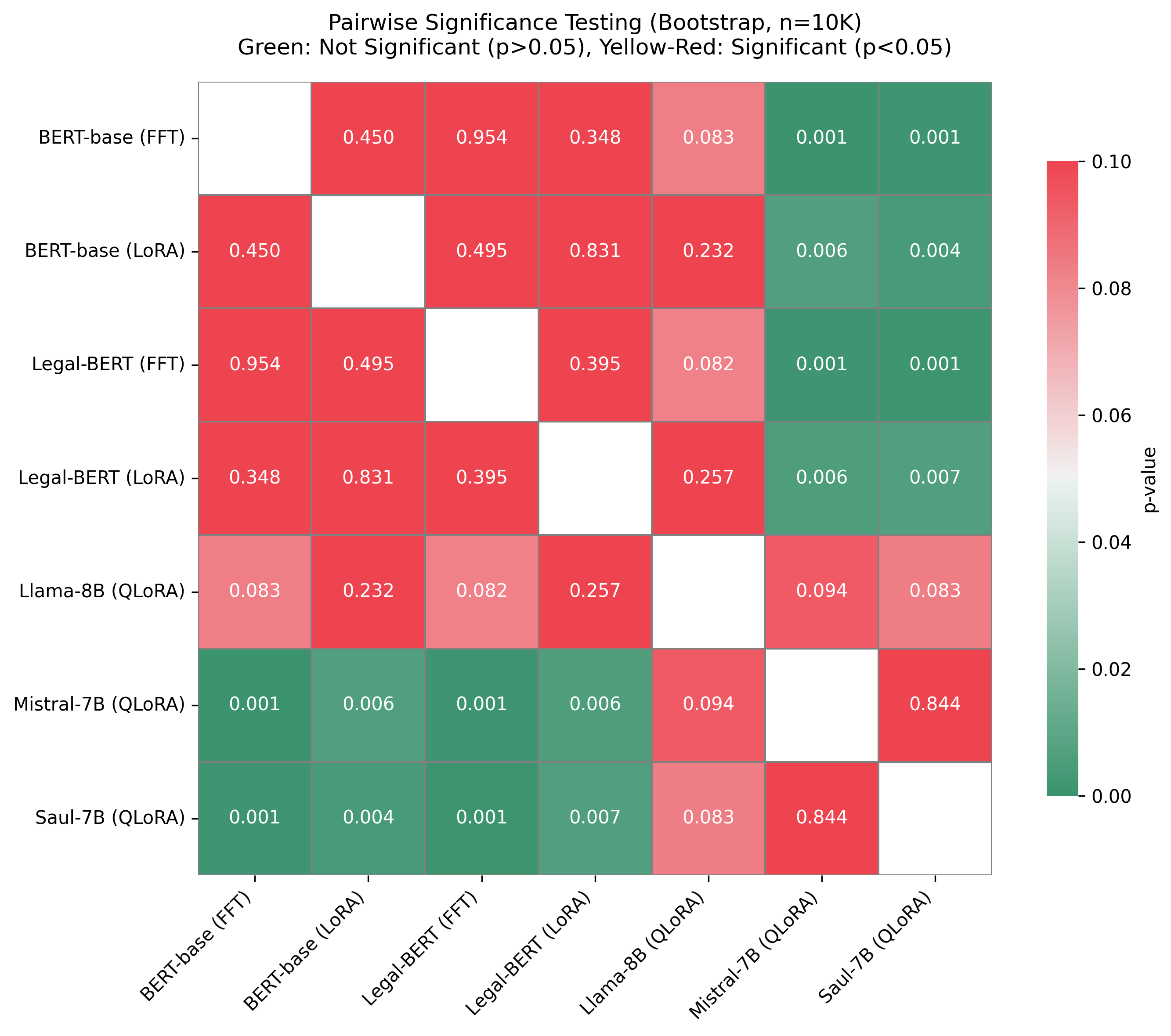}
\caption{Heatmap visualization of pairwise significance testing. Green indicates statistical equivalence ($p>0.05$), red indicates significant differences ($p<0.05$). Diagonal masked (self-comparisons). Color intensity reflects p-value magnitude.}
\label{fig:significance_heatmap}
\end{figure}

Table~\ref{tab:significance_matrix_full} presents all 21 unique pairwise comparisons (excluding self-comparisons along the diagonal), while Figure~\ref{fig:significance_heatmap} visualizes these relationships as a heatmap.

\begin{table}[h!]
\centering
\small
\caption{Complete 7$\times$7 pairwise significance testing. Upper triangle shows p-values for model comparisons. All tests use bootstrap paired testing ($n$=10K) except where noted.}
\label{tab:significance_matrix_full}
\setlength{\tabcolsep}{2pt}  
\renewcommand{\arraystretch}{1.1} 
\begin{tabular}{@{} l *{7}{c} @{}}
\toprule
\textbf{Model} & \textbf{B-FFT} & \textbf{B-LoRA} & \textbf{L-FFT} & \textbf{L-LoRA} & \textbf{Llama} & \textbf{Mistral} & \textbf{Saul} \\
\midrule
BERT-base FFT     & -- & 0.450 & 0.954 & 0.348 & 0.083 & \textbf{<0.001$^{\ddag}$} & \textbf{0.001$^{\ddag}$} \\
BERT-base LoRA    &    & --    & 0.495 & 0.831 & 0.232 & \textbf{0.006$^{\dag}$}  & \textbf{0.004$^{\dag}$} \\
Legal-BERT FFT    &    &       & --    & 0.395 & 0.082 & \textbf{0.001$^{\dag}$}  & \textbf{<0.001$^{\ddag}$} \\
Legal-BERT LoRA   &    &       &       & --    & 0.257 & \textbf{0.006$^{\dag}$}  & \textbf{0.007$^{\dag}$} \\
Llama-8B          &    &       &       &       & --    & 0.094 & 0.083 \\
Mistral-7B        &    &       &       &       &       & --    & 0.844 \\
Saul-7B           &    &       &       &       &       &       & -- \\
\bottomrule
\multicolumn{8}{p{0.95\linewidth}}{\scriptsize \textbf{Abbreviations:} B-FFT: BERT-base FFT; B-LoRA: BERT-base LoRA; L-FFT: Legal-BERT FFT; L-LoRA: Legal-BERT LoRA.} \\
\multicolumn{8}{p{0.95\linewidth}}{\scriptsize $^{*}p<0.05$, $^{\dag}p<0.01$, $^{\ddag}p<0.001$. BERT-to-BERT comparisons utilize Welch's t-test ($n$=5 seeds).} \\
\end{tabular}
\end{table}

\subsection{Interpretation}

The complete pairwise analysis reveals three statistically distinct performance clusters:

\begin{enumerate}[leftmargin=*, itemsep=1em]

\item \textbf{High-Performance Equivalence (Top Tier).} BERT-base FFT, Legal-BERT FFT, and Llama-8B form a statistically equivalent top tier (all pairwise p>0.08). This confirms that: (1) domain adaptation provides no significant advantage with optimal hyperparameters (BERT-base vs Legal-BERT: p=0.95), and (2) generative and discriminative paradigms achieve statistical parity (Llama vs both BERTs: p=0.08). All comparisons between this cluster and Cluster 3 show highly significant differences (p<0.01), confirming clear performance separation.

\item \textbf{Parameter-Efficient Tier (Middle).} BERT LoRA models form a middle tier, performing significantly worse than FFT counterparts (p<0.01) but showing internal equivalence (BERT-base LoRA vs Legal-BERT LoRA: p=0.83). The 9-point FFT advantage validates the accuracy-efficiency trade-off inherent in parameter-efficient adaptation. Comparisons with adjacent clusters show marginal significance or statistical equivalence, indicating smooth performance transitions.

\item \textbf{7B Generative Models (Lower Tier).} Mistral and Saul demonstrate statistical equivalence to each other (p=0.84) but perform significantly worse than Cluster 1 (all p<0.01). This 5-point deficit from Llama demonstrates scale effects (8B vs 7B) beyond domain adaptation, as legal-tuned Saul shows no advantage over generic Mistral.

\end{enumerate}

\section{Few-Shot Prompting Specification}\label{appendix_d}

This appendix provides complete specification of the few-shot prompting baseline described in Section \ref{subsec:baseline}, enabling reproducibility and clarifying methodological choices.

\subsection{Example Selection}

Five labeled examples were randomly sampled once from the training set (seed=42) and fixed across all 788 test instances. The examples comprised 3 positive (reporting obligations) and 2 negative (non-obligations) instances, providing balanced representation. Examples were selected without optimization for diversity or difficulty, constituting a naive baseline. Table~\ref{tab:fewshot_examples} presents the 5 examples randomly sampled (seed=42) and fixed across all test instances.

\begin{table}[h!]
\centering
\caption{Five-shot examples used in prompting baseline (3 positive, 2 negative).}\label{tab:fewshot_examples}
\small
\begin{tabular}{@{} cl p{0.7\textwidth} @{}}
\toprule
\textbf{ID} & \textbf{Label} & \textbf{Sentence} \\
\midrule
2538 & POS & In the case referred to in point (a) of the first subparagraph, where the group supervisor decides, after consulting the college of supervisors, no longer to include the subsidiary in the group supervision it carries out, it shall immediately inform the supervisory authority concerned and the parent undertaking. \\
\addlinespace
2949 & POS & Notifications shall be made within three working days of the transaction date to the competent authority of that Member State. \\
\addlinespace
3769 & POS & Without prejudice to paragraph 7, the Commission shall inform the European Parliament of the overall outcome of the application of paragraphs 5 and 6, including compliance with or derogation from the minimum percentages set per objective in the relevant Specific Regulations. \\
\addlinespace
1533 & NEG & Subject to Article 37(7), any consideration paid by the asset management vehicle in respect of the assets, rights or liabilities acquired directly from the institution under resolution shall benefit the institution under resolution. \\
\addlinespace
1507 & NEG & The decision of the competent authority to agree, prohibit or restrict the financial support shall be immediately notified to: \\
\bottomrule
\end{tabular}
\end{table}

\subsection{System Prompt}

The complete system instruction provided to the model:


\begin{tcolorbox}[templatebox, breakable, title={LLM Evaluation Prompt}]
Model Configuration \\
\textbf{Model:} Llama-3.1-8B Instruct --- \textbf{Sampling:} Disabled \\
\textbf{Temperature:} T=0 (deterministic) --- \textbf{Max tokens:} 512 \\
\hrule
\medskip
You are a legal expert analyzing EU legislation.

\textbf{Task:} Extract any reporting obligations from the provided text.

\textbf{Definition:} A reporting obligation is a mandatory legal requirement for an entity (e.g. operator, Member State, or authority) to submit specific information to a regulatory or oversight authority for purposes of supervision, enforcement, or regulatory coordination.

\textbf{Not reporting obligations (exclude these):}
\begin{itemize}
    \item Applications or requests (e.g. ``submit an application'', ``apply for registration'')
    \item Preparatory documentation (e.g. ``submit a monitoring plan'', ``submit a methodology'')
    \item Financial or operational requests (e.g. ``submit payment claims'', ``request reimbursement'')
    \item Template/format-only rules (e.g. ``shall use the template when notifying'')
    \item Authority-to-public or authority-to-applicant notifications (e.g. ``shall publish the report'', ``shall inform the applicant of its decision'')
\end{itemize}

\textbf{Output Format:}
\begin{itemize}
    \item If there are reporting obligations, output the exact sentence(s) from the text that contain them (no rephrasing).
    \item If no reporting obligation is present, respond with exactly: None
\end{itemize}
\hrule
\medskip
Examples formatted as:

Example 1:  [Sentence text] → [Full sentence or 'None']

Query: [Test sentence] →
\end{tcolorbox}

\section{Fine-Tuned LLM Evaluation Specification}\label{appendix_e}

This appendix provides complete specification of the evaluation procedure for fine-tuned generative models (Llama-3.1-8B, Mistral-7B, Saul-7B).

\subsection{Task Formulation}

Fine-tuned LLMs are evaluated using the same instruction format used during training: conditional text generation where the model receives a legislative sentence and must output either the extracted reporting obligation or "None" if no obligation exists.

\subsection{System Instruction (Training and Evaluation)}

The same system instruction is used for both training and evaluation:

\begin{tcolorbox}[templatebox, breakable, title={Fine-Tuned LLM System Instruction}]

You are a legal expert analyzing EU legislation.

\textbf{Task:} Extract any reporting obligations from the provided text.

\textbf{Definition:} A reporting obligation is a mandatory legal requirement for an entity (e.g. operator, Member State, or authority) to submit specific information to a regulatory or oversight authority for purposes of supervision, enforcement, or regulatory coordination.

\textbf{Not reporting obligations (exclude these):}
\begin{itemize}
    \item Applications or requests (e.g. ``submit an application'')
    \item Preparatory documentation (e.g. ``submit a monitoring plan'')
    \item Financial requests (e.g. ``submit payment claims'')
    \item Template-only rules (e.g. ``shall use the template'')
    \item Authority-to-public notifications (e.g. ``shall publish the report'')
\end{itemize}

\textbf{Output Format:}
\begin{itemize}
    \item If obligation exists: Output the exact sentence(s) containing it
    \item If no obligation: Respond with exactly "None"
\end{itemize}

\end{tcolorbox}

\subsection*{Input Format}

Each test instance is structured as a conversation with three roles:

\begin{verbatim}
conversations: [
  {role: 'system', content: [System instruction above]},
  {role: 'user', content: "Text: [Legislative sentence]"},
  {role: 'assistant', content: [Model generates]}
]
\end{verbatim}

Model-specific chat templates produce the following formats:

\begin{tcolorbox}[templatebox, breakable, title=Llama-3.1 Format]
\begin{small}
\begin{verbatim}
<|begin_of_text|><|start_header_id|>system<|end_header_id|>
[System instruction]<|eot_id|>
<|start_header_id|>user<|end_header_id|>
Text: [sentence]<|eot_id|>
<|start_header_id|>assistant<|end_header_id|>
\end{verbatim}
\end{small}
\end{tcolorbox}

\begin{tcolorbox}[templatebox, breakable, title=Mistral/Saul Format]
\begin{small}
\begin{verbatim}
<s>[INST][System instruction]

Text: [sentence][/INST]
\end{verbatim}
\end{small}
\end{tcolorbox}

The model generates the continuation starting after the final marker using greedy decoding (temperature=0, sampling disabled) for deterministic outputs. Generation uses a maximum of 384 new tokens with batch size of 16 for efficiency. Padding is applied left-side to support batch processing. Model-specific EOS tokens terminate generation: Llama uses dual terminators (\texttt{<|end\_of\_text|>}, \texttt{<|eot\_id|>}) while Mistral and Saul use standard \texttt{</s>} token.

\subsection{Evaluation}

Model outputs are first parsed into binary predictions: if the model generates obligation text, the prediction is positive; if it outputs "None" or remains empty, the prediction is negative. These predictions are then compared with ground truth labels to compute confusion matrices (TP/FP/TN/FN) and calculate Precision, Recall, and F1 scores.

\end{appendices}

\section*{Acknowledgments}
AWS resources were provided by the National Infrastructures for Research and Technology GRNET and funded by the EU Recovery and Resiliency Facility.

\section*{Declarations}
\textbf{Competing interests}. The authors have no competing interests to declare that are relevant to the content
of this article.

\textbf{Data and Code Availability}. The EURO-5K dataset, training and evaluation code, and trained model checkpoints are available in the companion repository: \url{https://github.com/ntua-el21432/EURO-5K-paper-companion-repo/}. The interactive web application integrating model predictions, explainability visualisations, and RRMV-compliant RDF export is available at: \url{https://github.com/ntua-el21432/EURO-5K-Web-Application}.

\bibliography{rep-obl}


\end{document}